\documentclass[manuscript,screen,nonacm]{acmart}%

\AtBeginDocument{%
  \providecommand\BibTeX{{%
    \normalfont B\kern-0.5em{\scshape i\kern-0.25em b}\kern-0.8em\TeX}}}

\setcopyright{none}%

\usepackage{todonotes}

\usepackage{enumitem}

\usepackage{fontawesome}

\usepackage[framemethod=tikz]{mdframed}
\definecolor{mycolor}{rgb}{0.122, 0.435, 0.698}
\definecolor{myblue}{rgb}{0.122, 0.435, 0.698}
\definecolor{mygreen}{rgb}{0.125, 0.525, 0.220}
\definecolor{myyellow}{rgb}{0.588, 0.439, 0.000}
\definecolor{myviolet}{rgb}{0.71764706, 0.40784314, 0.63529412}
\definecolor{myred}{rgb}{0.647, 0.114, 0.165}
\newmdenv[innerlinewidth=0.5pt,roundcorner=4pt,innerleftmargin=6pt,
          innerrightmargin=6pt,innertopmargin=6pt,innerbottommargin=6pt,
          linecolor=mycolor,backgroundcolor=mycolor!25!white]{mybox}
\newmdenv[innerlinewidth=0.5pt,roundcorner=4pt,innerleftmargin=6pt,
          innerrightmargin=6pt,innertopmargin=6pt,innerbottommargin=6pt,
          linecolor=myblue,backgroundcolor=myblue!25!white]{mybluebox}
\newmdenv[innerlinewidth=0.5pt,roundcorner=4pt,innerleftmargin=6pt,
          innerrightmargin=6pt,innertopmargin=6pt,innerbottommargin=6pt,
          linecolor=mygreen,backgroundcolor=mygreen!25!white]{mygreenbox}
\newmdenv[innerlinewidth=0.5pt,roundcorner=4pt,innerleftmargin=6pt,
          innerrightmargin=6pt,innertopmargin=6pt,innerbottommargin=6pt,
          linecolor=myyellow,backgroundcolor=myyellow!25!white]{myyellowbox}
\newmdenv[innerlinewidth=0.5pt,roundcorner=4pt,innerleftmargin=6pt,
          innerrightmargin=6pt,innertopmargin=6pt,innerbottommargin=6pt,
          linecolor=myred,backgroundcolor=myred!25!white]{myredbox}
\newmdenv[innerlinewidth=0.5pt,roundcorner=4pt,innerleftmargin=6pt,
          innerrightmargin=6pt,innertopmargin=6pt,innerbottommargin=6pt,
          linecolor=myviolet,backgroundcolor=myviolet!25!white]{myvioletbox}

\usepackage[skins]{tcolorbox}%
\newtcbox{\mylib}[2]{%
  enhanced,nobeforeafter,tcbox raise base,boxrule=0.4pt,top=0mm,bottom=0mm,%
  right=0mm,left=4mm,arc=1pt,boxsep=2pt,before upper=\vphantom{\Large dlg},%
  colframe=#2!50!black,coltext=#2!25!black,colback=#2!10!white,%
  overlay={\begin{tcbclipinterior}\fill[#2!75!blue!50!white] (frame.south west)%
    rectangle node[text=black,font=\sffamily\bfseries\footnotesize,rotate=90] {#1} ([xshift=4mm]frame.north west);\end{tcbclipinterior}}%
}%
\robustify{\mylib}

\usepackage{amsmath}
\usepackage{dsfont}

\usepackage{caption}
\usepackage{subcaption}

\usepackage{tikz}
\ExplSyntaxOn
\makeatletter
\tl_set:Nn \l_tmpa_tl {pgfsys-dvips.def}
\tl_if_eq:NNT \l_tmpa_tl \pgfsysdriver
  {\cs_set:Npn\pgfsys@blend@mode#1{\special{ps:~/\tl_upper_case:n #1~.setblendmode}}}
\makeatother
\ExplSyntaxOff
\usetikzlibrary{fit}
\usetikzlibrary{calc}
\usepackage{nicematrix}

\NiceMatrixOptions{renew-dots,renew-matrix,%
                   create-medium-nodes,
                   left-margin,
                   right-margin}%
\tikzset{highlight_red/.style={rectangle,
                               fill = red!15,
                               blend mode = multiply,
                               name suffix = -medium,
                               rounded corners = 0.5 mm,
                               inner sep = 1pt,
                               fit = #1}}
\tikzset{highlight_blue/.style={rectangle,
                                fill = blue!15,
                                blend mode = multiply,
                                name suffix = -medium,
                                rounded corners = 0.5 mm,
                                inner sep = 1pt,
                                fit = #1}}

\newcommand{\RGB}[3]{(\parbox{\widthof{255}}{\hfill#1}, \parbox{\widthof{255}}{\hfill#2}, \parbox{\widthof{255}}{\hfill#3})}

\newcommand{\IR}{\ensuremath{\mathit{IR}}}

\graphicspath{{./fig/}}

\begin{document}

\title[%
Interpretable Representations in Explainable AI%
]{%
Interpretable Representations in Explainable AI: %
From Theory to Practice%
}

\author{Kacper Sokol}
\authornote{Corresponding author.}
\email{k.sokol@bristol.ac.uk}
\email{kacper.sokol@inf.ethz.ch}
\orcid{0000-0002-9869-5896}
\affiliation{%
  \institution{Intelligent Systems Laboratory, University of Bristol}
  \country{United Kingdom}
}
\affiliation{%
  \institution{Department of Computer Science, ETH Zurich}
  \country{Switzerland}
}

\author{Peter Flach}
\email{peter.flach@bristol.ac.uk}
\orcid{0000-0001-6857-5810}
\affiliation{%
  \institution{Intelligent Systems Laboratory, University of Bristol}
  \country{United Kingdom}
}

\begin{abstract}
Interpretable representations are the backbone of many explainers that target black-box predictive systems based on artificial intelligence and machine learning algorithms. %
They translate the low-level data representation necessary for good predictive performance into high-level human-intelligible concepts used to convey the explanatory insights. %
Notably, the explanation type and its cognitive complexity are directly controlled by the interpretable representation, tweaking which allows to target a particular audience and use case. %
However, many explainers built upon interpretable representations overlook their merit and fall back on default solutions that often carry implicit assumptions, thereby degrading the explanatory power and reliability of such techniques. %
To address this problem, we study properties of interpretable representations that encode presence and absence of human-comprehensible concepts. %
We demonstrate how they are operationalised for tabular, image and text data; discuss their assumptions, strengths and weaknesses; identify their core building blocks; and scrutinise their configuration and parameterisation. %
In particular, this in-depth analysis allows us to pinpoint their explanatory properties, desiderata and scope for (malicious) manipulation in the context of tabular data where a linear model is used to quantify the influence of interpretable concepts on a black-box prediction. %
Our findings lead to a range of recommendations for designing trustworthy interpretable representations; %
specifically, the benefits of class-aware (supervised) discretisation of tabular data, e.g., with decision trees, and sensitivity of image interpretable representations to segmentation granularity and occlusion colour.%
\end{abstract}

\keywords{%
Interpretability; %
Explainability; %
Surrogates; %
Post-hoc; %
Interpretable Representations; %
Machine Learning; %
Artificial Intelligence.%
}%

\maketitle

\vfill
\begin{mygreenbox}
  \textbf{Highlights}%
  \begin{itemize}[topsep=0pt,label=\faLightbulbO,leftmargin=.5cm,itemindent=0cm,labelwidth=.5cm,labelsep=0cm,align=left]%
      \item Interpretable representations need to be crafted for the problem at hand -- ideally with a human in the loop -- to become a trustworthy foundation of explainability.%
      \item The information removal proxy required by image and tabular interpretable representations should be deterministic and domain-aware.%
      \item For images, the occlusion colour and segmentation granularity play important roles, with mean-colour occlusion exhibiting a number of undesired properties.%
      \item For tabular data, fidelity of continuous feature discretisation is critical, with class-aware methods yielding best results.%
      \item Tabular data explainers built upon discretisation-based interpretable representations combined with surrogate linear models are fragile, possibly misleading, and can be easily manipulated due to information loss.%
  \end{itemize}
\end{mygreenbox}
\vspace{.33em}%
\begin{mybluebox}
\noindent\faGithub\hspace{.2cm}\textbf{Source Code}\quad%
\url{https://github.com/So-Cool/bLIMEy/tree/master/DAMI_2024}
\end{mybluebox}
\vspace{.33em}%
\begin{myvioletbox}
\noindent\faFileTextO\hspace{.2cm}\textbf{Published in}\quad%
Data Mining and Knowledge Discovery (\href{https://doi.org/10.1007/s10618-024-01010-5}{10.1007/s10618-024-01010-5})%
\end{myvioletbox}
\vfill

\clearpage\newpage

\section{Introduction}
Interpretable Representations (IRs) are the foundation of many explainability methods that target black-box predictive models based on Artificial Intelligence (AI) and Machine Learning (ML) algorithms. %
They are the core component of surrogate explainers, which are techniques to approximate the functioning and behaviour of an unintelligible classifier or regressor with a simpler model~\cite{friedman2008predictive,ribeiro2016why,lundberg2017unified,sokol2019blimey}. %
More broadly, IRs facilitate translating the ``language'' of ML models -- low-level data representations required for good predictive performance, such as raw feature values and their complex embeddings -- into high-level concepts that are understandable to humans. %
IRs, therefore, create an \emph{interface} between a computer-readable encoding of a phenomenon (captured by the collected data) and cognitively digestible chunks of information, thus establishing a medium suitable for conveying explanations. %
Notably, interpretable representations directly control the (perceived) complexity of the ensuing explanations, define the question that these insights answer, and restrict the explanation types that can effectively communicate this information -- e.g., influence or importance of interpretable concepts, counterfactuals and what-if statements -- making IR-based explainers highly flexible, versatile and appealing. %
In essence, by customising the interpretable representation we can adjust the content and comprehensibility of the resulting explanations and tune them towards a particular \emph{audience} and \emph{application}.%

The algorithmic process responsible for transforming data from their original domain into an interpretable representation is usually defined by a human. %
An IR of images, for example, can be created with a \emph{super-pixel segmentation}, i.e., partitioning images into non-overlapping clusters of pixels, each one representing an object of interest or pieces thereof. %
Similarly, text can be split into \emph{tokens} denoting individual words, their stems or collections of words that are not necessarily adjacent. %
Tabular data containing numerical features can be \emph{discretised} to capture meaningful patterns, e.g., people belonging to different age groups. %
Such interpretable representations are often paired with a simple and inherently transparent model to form a \emph{surrogate explainer}; for example, LIME -- Local Interpretable Model-agnostic Explanations~\cite{ribeiro2016why} -- uses a sparse linear model. %
IR-based explainers are thus data-universal, in addition to often being model-agnostic and post-hoc~\cite{sokol2020explainability}, which makes them an attractive choice given that they can be used with pre-existing black-box ML models. %

Given the high complexity of such end-to-end explainers, many of them are composed of generic, thus versatile, algorithmic building blocks and focus on maximising their overall performance, hence forgoing selection and optimisation of their individual parts~\cite{sokol2019blimey}. %
Moreover, these explainers seek to automate the entire process to enable their deployment and evaluation at scale, which understandably requires components that can be operationalised without human input. %
Given the vast array of algorithmic choices in this space -- as well as their individual configurations -- such explainers are in fact complex entities suffering from overparameterisation, which often manifests itself in multiple contributing sources of randomness and low fidelity of the resulting explanations~\cite{zhang2019should,rudin2019stop,lakkaraju2019faithful,lakkaraju2020fool,sokol2019blimey,sokol2020limetree}. %
This observation is particularly pertinent to interpretable representations, popular examples of which are: \emph{quantile discretisation} for numerical features of tabular data; edge-based \emph{super-pixel segmentation} for images, e.g., via quick shift~\cite{vedaldi2008quick}; and whitespace-based \emph{tokenisation} for text~\cite{ribeiro2016why}.%

Many deficiencies plaguing such explainers can be attributed to misuse of the underlying interpretable representation, which can make or break an explainer~\cite{sokol2019blimey,sokol2020limetree,sokol2021towards}. %
These problems can be magnified, possibly rendering the explainer unusable, through certain pairings of IRs and types of surrogate models, especially when the implicit assumptions behind both of these components are at odds. %
By understanding the characteristics and behaviour of each interpretable representation and its influence on the resulting explanations -- both on its own and in conjunction with a particular surrogate model family -- we can explicate the theoretical properties of such explainers and assess their applicability and usefulness for the problem at hand. %
This area of research is largely under-explored for IRs on their own and as a part of an explainer, potentially leading to suboptimal design choices and inadequate explanations. %
An especially impactful research direction, which we focus on in this paper, is \emph{automatic} creation of IRs that are trustworthy, robust and faithful to enable their creation, optimisation and deployment with minimal human input.%

The important tasks of choosing an appropriate interpretable representation and configuring it are not often considered in the literature. %
It is common to assume that an IR is given or to reuse one that was proposed in prior work without much afterthought or deliberation about its suitability, (implicit) assumptions, properties and caveats~\cite{laugel2018defining,zhang2019should,lakkaraju2020fool}. %
As a result the interpretable representations introduced along the first explainers utilising them are still dominating the explainability landscape and are widely used despite possibly being a subpar choice. %
Specifically, the most popular use case of IRs -- in which they are deployed to measure the positive or negative \emph{influence} of each interpretable component (more precisely, information that it encodes) on a black-box prediction of a selected instance -- comes with many unaddressed issues. %
For example, to carry out this \emph{sensitivity analysis}, a random subset of IR elements needs to be ``removed'' a number of times and the resulting change in the model's prediction quantified, e.g., by the coefficients of a (surrogate) linear model. %
Most black-box models, however, cannot predict incomplete instances, especially for tabular and image data, in which case this procedure becomes ill-defined and replaced with a proxy operation -- such as segment occlusion for images -- potentially leading to biased and untrustworthy explanations.%

In this paper, we investigate the capabilities and limitations of the most common type of interpretable representations, where presence and absence of interpretable concepts is encoded with a binary on/off vector. %
We first overview relevant literature and introduce popular IRs for text, image and tabular data in Section~\ref{sec:ir}, where we also show example explanations built upon them. %
This section additionally identifies the core elements, parameterisation and deficiencies of interpretable representations, which facilitates their analytical and experimental investigation. %
Next, in Section~\ref{sec:ir-analysis}, we study the influence of suboptimal configuration of IRs and the implications of employing various algorithmic proxies necessary to make them computationally feasible and scalable whenever it is impossible or impractical to directly remove information from the underlying data. %
We also investigate implicit assumptions such as the \emph{locality} of an explanation, which may prevent its completeness, and the \emph{stochasticity} of the transformation between the original and interpretable domains (and vice versa), which may introduce unnecessary randomness, contribute to volatility and reduce fidelity and soundness of explanations, thereby harming their \emph{veracity}~\cite{sokol2019blimey,sokol2020explainability}. %
Our findings are supported by a range of experiments that analyse these factors for quartile-based discretisation and decision tree-based partition of numerical features for tabular data -- where information removal is achieved through a random allocation of a feature value to one of these attribute ranges -- as well as super-pixel segmentation of images with varying granularity -- for which colour-based occlusion is used as an information removal strategy.%

In Section~\ref{sec:ols} we examine the lineage and interpretation of influence-based explanations -- determined by the coefficients of a linear model in a surrogate explainer setting -- with respect to the properties of the underlying interpretable representation for tabular data with numerical features. %
In particular, we illustrate the limited explanatory capabilities of an IR built upon (unsupervised) discretisation of continuous attributes when paired with Ordinary Least Squares (OLS), i.e., a bare-bones version of LIME, for which an analytical (closed-form) solution is derived in Appendix~\ref{sec:appendix:ols}. %
Such explainers can lose the precise encoding of the black-box decision boundary and be (externally) manipulated by altering the distribution of the data sample used to fit the surrogate OLS, both of which undermine reliability of the ensuing explanations. %
As a solution we propose using \emph{supervised} discretisation algorithms that produce up to \emph{three bins} per numerical feature in addition to employing alternative types of surrogate models -- a recommendation supported by a collection of theoretical and experimental results. %
Specifically, we investigate decision trees, which prove to be particularly suitable in this setting since they can both partition (i.e., discretise) the data space to crate meaningful interpretable concepts, and generate a wide array of appealing explanations such as exemplars, importance scores and counterfactuals~\cite{waa2018contrastive,sokol2021towards}.%

All of these findings allow us to create guidelines for building trustworthy, faithful and algorithmically sound interpretable representations, which we outline in Section~\ref{sec:guidelines}. %
This collection of insights is a stepping stone towards automatic generation of robust IRs with well-known properties and caveats. %
It also highlights how state-of-the-art research in fields such as natural language processing, image segmentation and discretisation of tabular data can inform better design of interpretable representations, in addition to discussing the beneficial role of humans in this process. %
Furthermore, our results demonstrate the importance of developing representative validation criteria and metrics for individual components of explainability algorithms, which is an improvement over evaluating only the final, end-to-end explainer~\cite{sokol2020tut,small2023helpful,xuan2023can,sokol2024what}. %
We conclude this paper with Section~\ref{sec:conclusion}, which summarises our findings and discusses future research directions.%

\section{Interpretable Representations\label{sec:ir}}%

The choice of the interpretable representations and surrogate models used for our analysis is motivated by the popularity of these individual components in the literature. %
Specifically, LIME~\cite{ribeiro2016why} and RuleFit~\cite{friedman2008predictive} use a surrogate \emph{linear} model to estimate influence of interpretable concepts. %
Additionally, LIME and SHAP -- SHapley Additive exPlanations~\cite{lundberg2017unified} -- employ an interpretable representation that encodes \emph{presence and absence} of intelligible concepts to formulate their explanations. %
Similarly, \citet{friedman2008predictive} experimented with automatic learning of more complex IRs for tabular data by fitting a random forest and extracting rules from it. %
These logical statements, which capture various concepts, are then used as binary meta-features to train a linear model, thus offering a highly expressive interpretable representation. %
More recently, \citet{garreau2020explaining} analysed theoretical properties and parameterisation of vanilla LIME for tabular data, including its interpretable representation and surrogate linear model; however, their work treated the explainer as an end-to-end algorithm and operated under quite restrictive assumptions, e.g., presupposing linearity of the underlying black box.%

We focus on interpretable representations of tabular, image and text data given their dominant role in XAI. %
While the operationalisation of IRs varies across these data types, their machine representation is usually consistent: a binary vector indicating presence (\emph{fact} denoted by \(1\)) or absence (\emph{foil} denoted by \(0\)) of certain human-understandable concepts for a selected data point. %
The IRs of image and text data are relatively intuitive and share many properties. %
Images are partitioned into non-overlapping segments called super-pixels, which are then represented in the interpretable binary space as either preserved (i.e., original pixel values) or removed. %
Similarly, text is split into tokens that can encode individual words, their stems or collections of words, the presence or absence of which is captured by the IR. %
Tabular data, on the other hand, are more problematic since, first, numerical attributes need to be discretised to create a hyper-rectangle partition of the feature space, followed by a binarisation procedure that for each (now discrete) dimension records whether a data point is located within or outside of the hyper-rectangle selected to be explained.%

The interpretable representations of text and images are reasonably easy to generate automatically and (when configured correctly) the meaning of the resulting explanations is relatively accessible to a lay audience -- a characteristic that is not necessarily true of tabular data as we will see later. %
Additionally, the high dimensionality of raw text and image data does not impair their comprehensibility, but it does for tabular data as humans are generally confined to three dimensions given the inherent spatio-temporal limitations of our visual apparatus. %
Dimensionality reduction for images and text is thus unnecessary and may even be harmful; removal of super-pixels is an ill-defined procedure that results in blank spaces, whereas discarding stop words and punctuation marks as well as word transformations can be considered as pruning steps that should be incorporated directly into the interpretable representation composition function and executed prior to tokenisation. %
The process of transforming data from their original domain into an interpretable representation is in most cases defined by the user and built into the (surrogate) explainer. %
Uniquely for tabular data, however, it can be learnt as part of the \emph{explanation generation} step depending on the surrogate model choice~\cite{sokol2019blimey,sokol2020limetree,sokol2021towards}. %
Notably, specifying the foil -- i.e., the operation linked to switching off a component of the IR by setting its binary value to \(0\) -- may not always be straightforward or even feasible in certain domains, requiring a problem-specific information removal proxy~\cite{mittelstadt2019explaining}.%

\subsection{Text\label{sec:ir:text}}%
The interpretable domain based on presence and absence of tokens in text feels natural and appealing to humans. %
Individual words and their groups encode understandable concepts and their absence may alter the meaning of a sentence, which arguably reflects how humans comprehend text. %
A na\"ive IR can represent text as a bag of words, where each word becomes a token, thereby forgoing the influence of word ordering and the information carried by their co-occurrence. %
We can easily improve upon that and capture the dependencies between words by including \(n\)-gram groupings. %
Applying other pre-processing steps, e.g., extracting word stems or lemmatisation, can also be beneficial for the human-comprehensibility of such interpretable representations. %
Machine processing of (natural language) text is a well-established research field~\cite{manning1999foundations} that can be a rich source of inspiration for designing appealing and informative IRs.%

\begin{figure}[b]
  \centering
  \begin{subfigure}[t]{.99\textwidth}
      \centering
      \includegraphics[height=1.625cm]{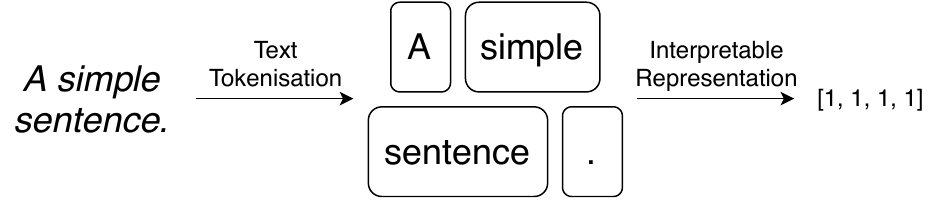}%
      \caption{Transformation from the original domain into the interpretable representation \(\mathcal{X} \rightarrow \mathcal{X}^\star\).\label{fig:txt:1}}%
  \end{subfigure}
  \par\bigskip %
  \begin{subfigure}[t]{.99\textwidth}
      \centering
      \includegraphics[height=1.625cm]{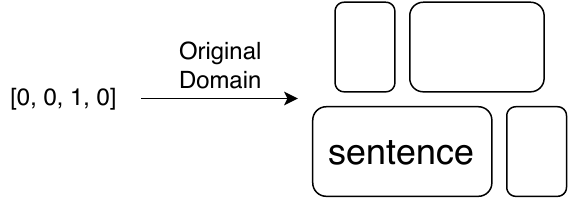}%
      \caption{Transformation from the interpretable representation into the original domain \(\mathcal{X}^\star \rightarrow \mathcal{X}\).\label{fig:txt:2}}%
  \end{subfigure}
  \caption{%
Depiction of a forward and backward transformation between the original and interpretable representations of text data. %
Panel~(\subref{fig:txt:1}) shows steps required to represent a sentence as a binary on/off vector; Panel~(\subref{fig:txt:2}) illustrates this procedure in the opposite direction. %
Both transformations are \emph{deterministic} given a fixed algorithm responsible for text pre-processing and tokenisation.%
\label{fig:txt}}%
\end{figure}

\begin{figure}[t]%
  \centering
  \begin{subfigure}[t]{0.4\textwidth}%
    \centering
    \parbox[b][\textwidth][c]{\textwidth}{
    \noindent\begin{flushleft}\mylib{\(x^\star_1\)}{white}{\Large This} \mylib{\(x^\star_2\)}{green}{\Large sentence} \mylib{\(x^\star_3\)}{white}{\Large has} \mylib{\(x^\star_4\)}{white}{\Large a} \mylib{\(x^\star_5\)}{green}{\Large positive} \mylib{\(x^\star_6\)}{red}{\Large sentiment} \mylib{\(x^\star_7\)}{white}{\Large ,} \mylib{\(x^\star_8\)}{red}{\Large maybe} \mylib{\(x^\star_9\)}{white}{\Large .}\end{flushleft}%
    }
    \caption{Words and punctuation marks are the components of the interpretable representation.\label{fig:text_ex:1}}
  \end{subfigure}
  \hspace{0.066666667\linewidth}
  \begin{subfigure}[t]{0.4\textwidth}%
    \centering
    \includegraphics[width=.75\textwidth]{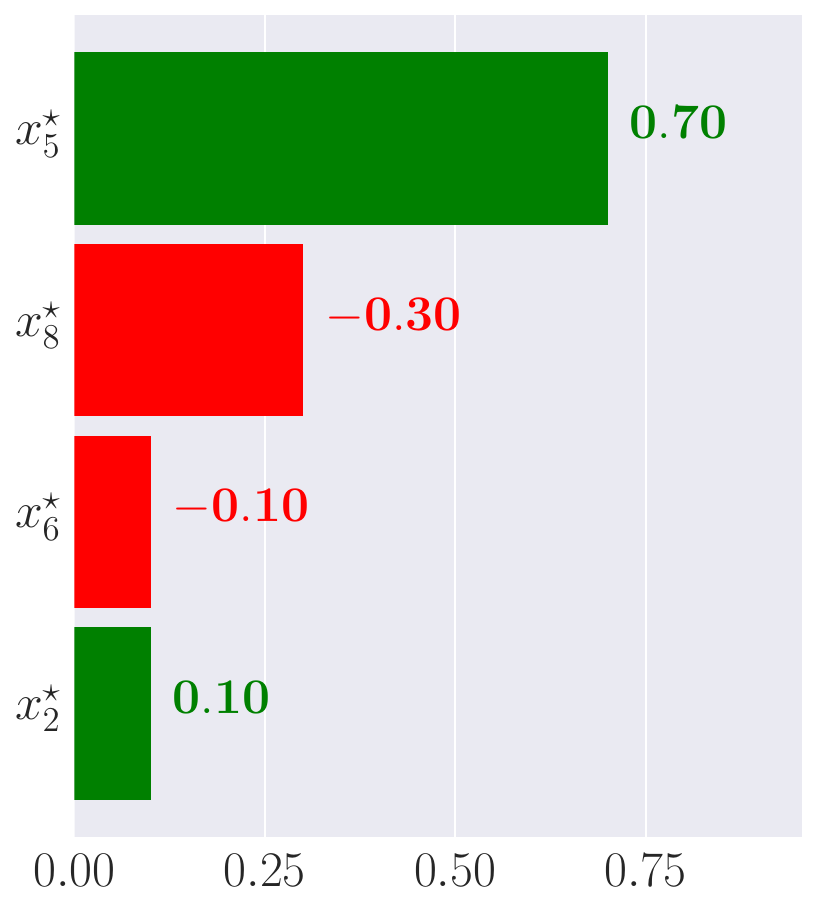}%
    \caption{Explanations capture influence of tokens when classifying the sentiment of a text excerpt.\label{fig:text_ex:2}}%
  \end{subfigure}
  \caption{Example of an influence-based explanation of text with a \emph{bag-of-words} interpretable representation. Panel~(\subref{fig:text_ex:1}) illustrates a sentence whose (positive) \emph{sentiment} is being decided by a black-box model. The colouring of each token in Panel~(\subref{fig:text_ex:1}) conveys its influence on the prediction, with Panel~(\subref{fig:text_ex:2}) depicting their respective magnitudes.\label{fig:text_ex}}%
\end{figure}

Once text is pre-processed and tokenised, it is \emph{deterministically} transformed into the binary interpretable representation. %
To this end, a sentence is encoded as a Boolean vector of length equal to the number of tokens in the IR, where \(1\) indicates presence of a given token and \(0\) its absence -- see Figure~\ref{fig:txt} for a demonstration of this procedure. %
The original sentence is thus encoded with an all-\(1\) vector. %
By flipping some components of this vector to \(0\), we effectively remove tokens from the underlying sentence and create its variations. %
Notably, the high dimensionality of this representation does not affect the readability of the resulting explanations since altered text cannot have more tokens than the original sentence. %
Explanations based on token influence can be overlaid on top of text by highlighting each token with a different shade of green (for positive influence) or red (for negative influence), thus expressing their respective influence on the explained class -- see Figure~\ref{fig:text_ex} for an example.%

\subsection{Images\label{sec:ir:images}}%

\begin{figure}[t]
  \centering
  \begin{subfigure}[t]{.99\textwidth}
      \centering
      \includegraphics[height=2.8125cm]{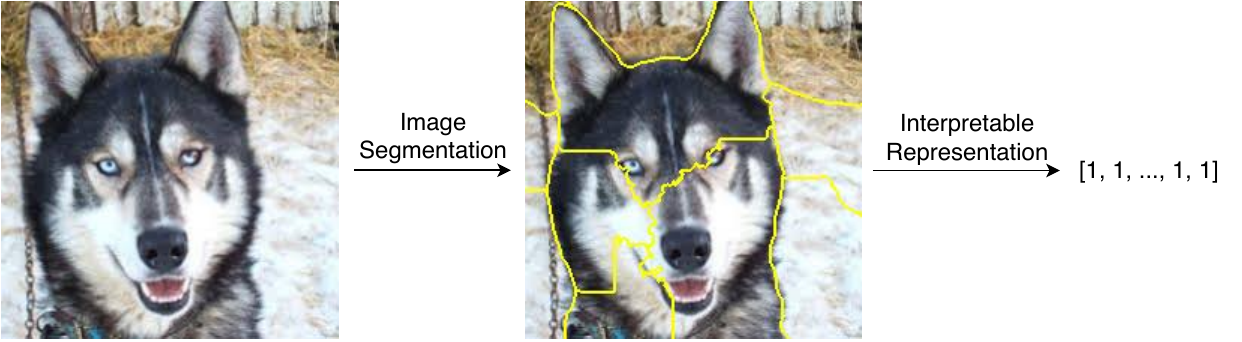}%
      \caption{Transformation from the original domain into the interpretable representation \(\mathcal{X} \rightarrow \mathcal{X}^\star\).\label{fig:img:1}}%
  \end{subfigure}
  \par\bigskip %
  \begin{subfigure}[t]{.99\textwidth}
      \centering
      \includegraphics[height=2.8125cm]{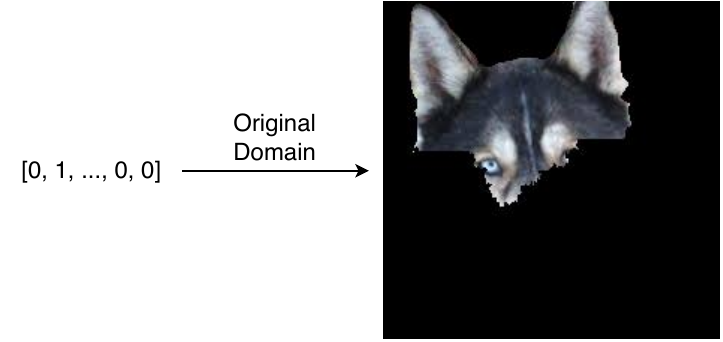}%
      \caption{Transformation from the interpretable representation into the original domain \(\mathcal{X}^\star \rightarrow \mathcal{X}\).\label{fig:img:2}}%
  \end{subfigure}
  \caption{%
Depiction of a forward and backward transformation between the original and interpretable representations of image data. %
Panel~(\subref{fig:img:1}) shows steps required to represent a picture as a binary on/off vector; Panel~(\subref{fig:img:2}) illustrates this procedure in the opposite direction. %
Both transformations are \emph{deterministic} given fixed image segmentation and occlusion strategies.\label{fig:img}}%
\end{figure}

\begin{figure}[t]%
    \centering
    \begin{subfigure}[t]{0.4\textwidth}%
        \centering
        \includegraphics[width=.75\textwidth]{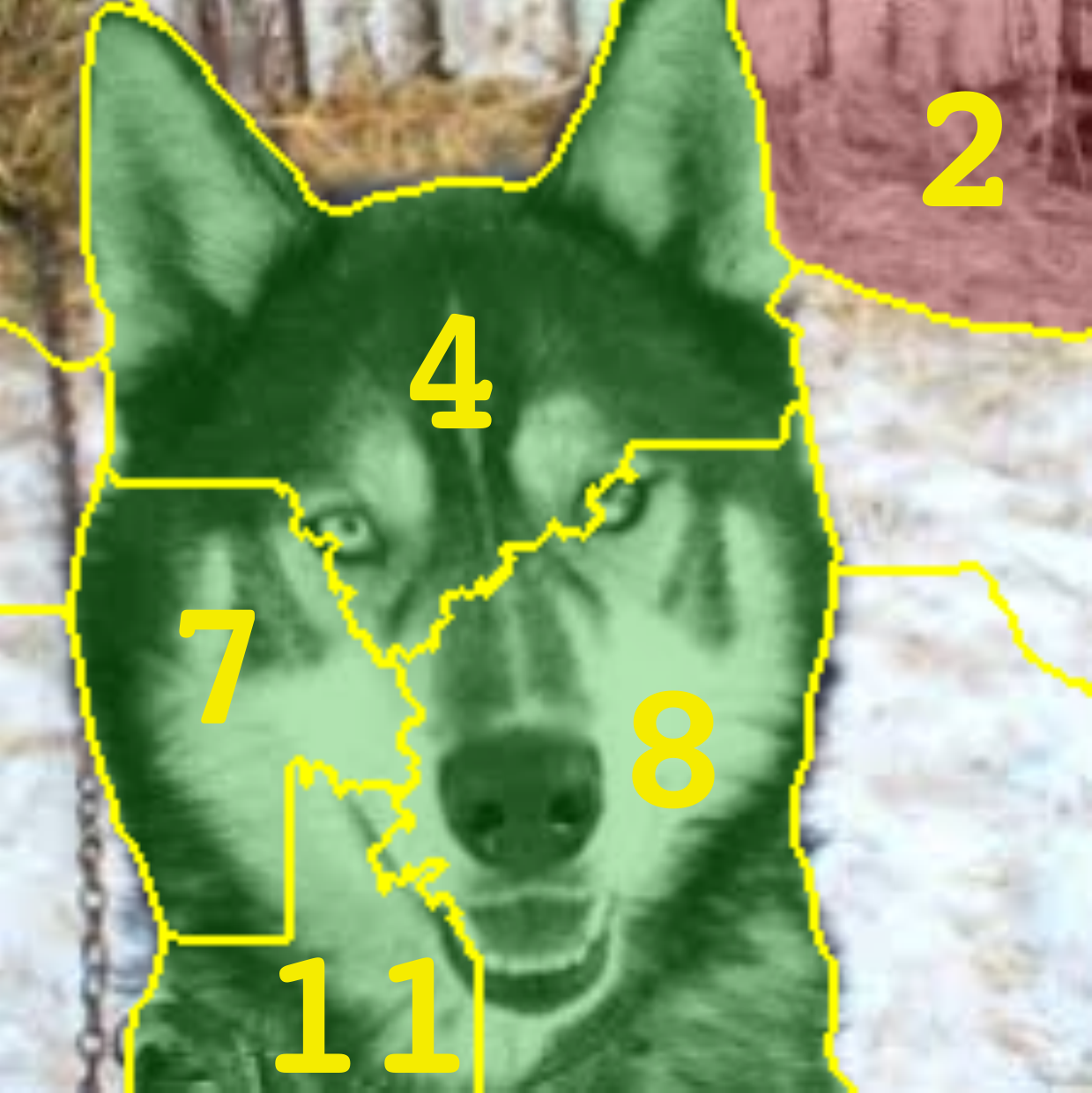}%
        \caption{Image segments are the components of the interpretable representation.\label{fig:img_ex:1}}
    \end{subfigure}
    \hspace{0.066666667\linewidth}
    \begin{subfigure}[t]{0.4\textwidth}%
        \centering
        \includegraphics[width=.75\textwidth]{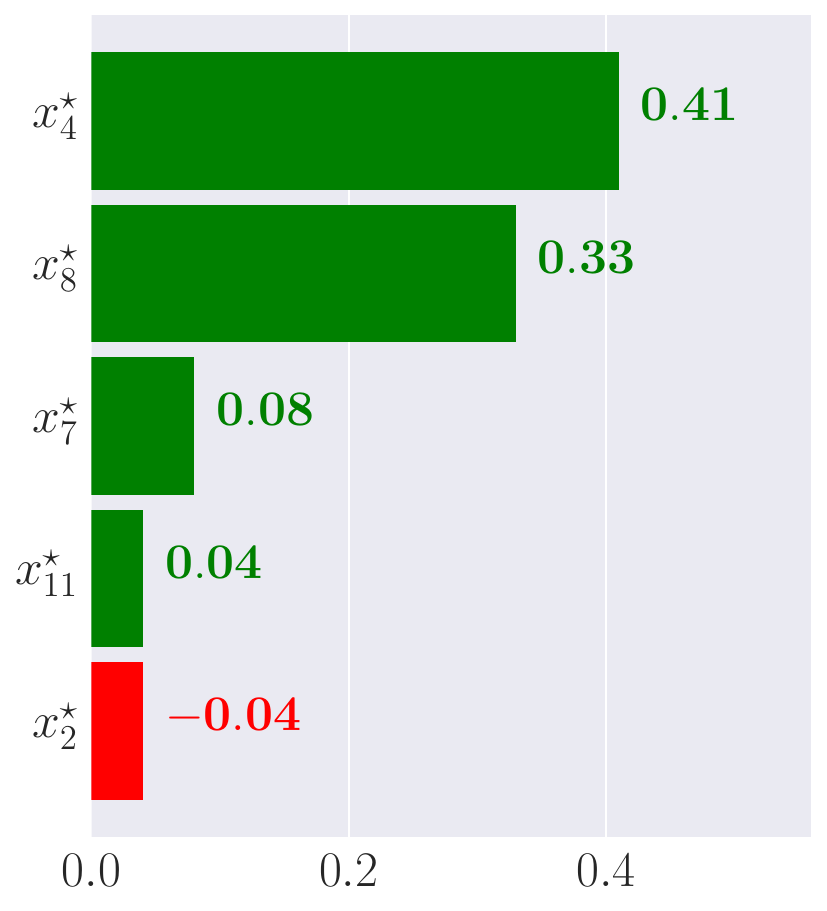}%
        \caption{Explanations show influence of the super-pixels on the selected class prediction (\emph{Eskimo dog}).\label{fig:img_ex:2}}%
    \end{subfigure}
    \caption{Example of an influence-based explanation of image data with the interpretable representation built upon \emph{segmentation}. Panel~(\subref{fig:img_ex:1}) illustrates an image that is being classified by a black-box model. The colouring of each super-pixel in Panel~(\subref{fig:img_ex:1}) conveys its influence on the prediction of a user-selected class (\emph{Eskimo dog} in this case), with Panel~(\subref{fig:img_ex:2}) depicting their respective magnitudes.\label{fig:img_ex}}%
\end{figure}

Interpretable representations of image data rely on the same premise: images are algorithmically segmented into super-pixels, often using edge-based methods such as quick shift~\cite{vedaldi2008quick,ribeiro2016why}. %
The presence (\(1\)) or absence (\(0\)) of these segments is manipulated by the underlying binary representation, where an all-\(1\) vector corresponds to the original picture -- see Figure~\ref{fig:img} for a reference. %
However, since a segment of an image cannot be directly removed given that relevant classifiers are unable to handle missing data -- in contrast to the equivalent procedure for text IRs -- setting one of the interpretable components to \(0\) is an ill-defined operation. %
Instead, a \emph{computationally-feasible proxy} is commonly used to hide or discard the information carried by super-pixels; specifically, segments are occluded with a solid colour. %
For example, LIME uses the mean colour of each super-pixel to mask its content~\cite{ribeiro2016why}. %
Explanations based on such interpretable representations communicate the influence of each image segment on the black-box prediction of a user-selected class as shown in Figure~\ref{fig:img_ex}.%

This approach comes with its own implicit assumptions and limitations. %
For one, an edge-based partition of an image may not capture concepts that are meaningful from a human perspective. %
\emph{Semantic segmentation} or outsourcing this task to the user appears to yield better results~\cite{sokol2020limetree,sokol2020one}, possibly at the expense of automation difficulties. %
Furthermore, the information removal proxy could be improved by replacing colour-based occlusion of super-pixels with a more meaningful process that better reflects how humans perceive visual differences in images. %
For example, the content of a segment could be occluded with another object, akin to the procedure proposed by Benchmarking Attribution Methods~\cite{yang2019bam}, or retouched in a context-aware manner, e.g., inpainted with what is anticipated in the background, thus preserving the integrity and colour continuity of the explained image. %
While both of these approaches are intuitive, they are difficult to automate and scale since the underlying operations are mostly limited to image partitions where each super-pixel represents a self-contained and semantically coherent object.%

\begin{figure}[b]
  \centering
  \begin{subfigure}[t]{.99\textwidth}
      \centering
      \includegraphics[height=2.5cm]{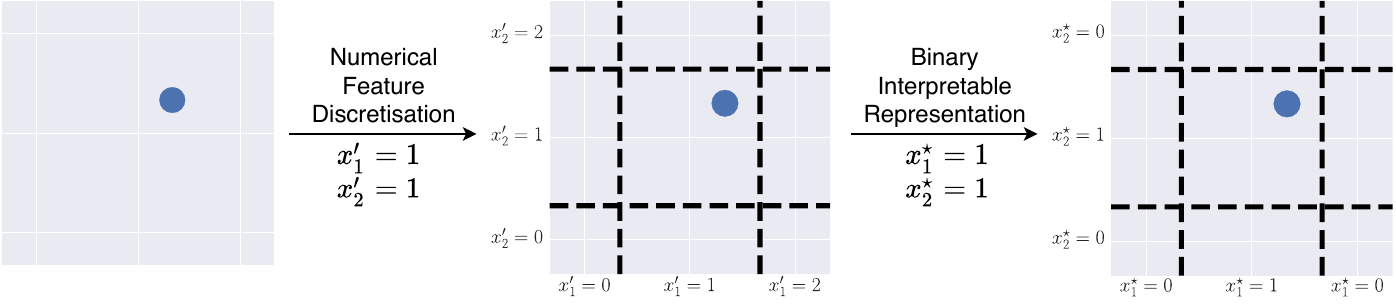}%
      \caption{Transformation from the original domain into the interpretable representation \(\mathcal{X} \rightarrow \mathcal{X}^\star\).\label{fig:tab:1}}%
  \end{subfigure}
  \par\bigskip %
  \begin{subfigure}[t]{.99\textwidth}
      \centering
      \includegraphics[height=2.5cm]{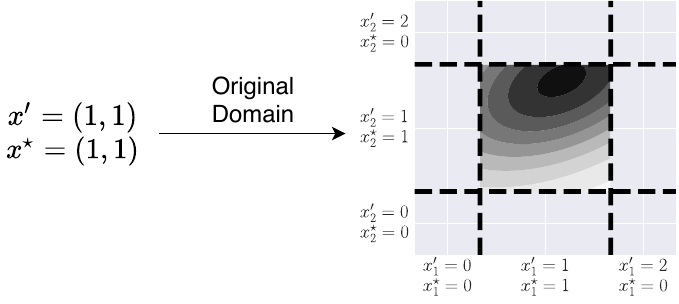}%
      \caption{Transformation from the interpretable representation into the original domain \(\mathcal{X}^\star \rightarrow \mathcal{X}\).\label{fig:tab:2}}%
  \end{subfigure}
  \caption{%
Depiction of a forward and backward transformation between the original and interpretable representations of tabular data. %
Panel~(\subref{fig:tab:1}) shows the discretisation and binarisation steps required to represent a data point as a binary on/off vector; Panel~(\subref{fig:tab:2}) illustrates this procedure in the opposite direction. %
The forward transformation is \emph{deterministic} given a fixed discretisation algorithm (i.e., binning of numerical features); however, moving from the IR to the original domain is \emph{stochastic} since it requires random sampling.%
\label{fig:tab}}%
\end{figure}

\subsection{Tabular Data\label{sec:ir:tabular}}%

The raw features used by predictive models trained on images and text -- e.g., pixel values and word embeddings -- are often difficult to reason about, establishing the need for interpretable representations. %
In contrast, tabular data may not require an IR to become explainable if their attributes are human-comprehensible from the outset. %
On the other hand, if the explanation is to answer a specific question -- as is the case for images and text -- using an interpretable representation may be helpful. %
Continuing with the theme of investigating concept influence, for tabular data we are interested in how presence and absence of certain \emph{binary} characteristics, which the explained data point exhibits, affect its prediction. %

One approach is to treat the specific attribute values of the explained instance as concepts: if a feature value of a data point is identical to the value of the same attribute in the explained instance, the concept is \emph{present} (\(1\)), otherwise it is \emph{absent} (\(0\)), which procedure becomes the information removal proxy. %
For example, if the second feature \(x_2\) of the explained instance \(\mathring{x}\) is \(\mathring{x}_2 = 70\), any instance \(x\) whose second attribute has the same value (\(70\)) is assigned \(x^\star_2 = 1\) in the binary IR, and \(0\) otherwise, i.e.,%
\[
x^\star_i =
\begin{cases}
    1    & \text{if~} x_i = \mathring{x}_i \text{,} \\
    0    & \text{otherwise.}
\end{cases}
\]
While this may be appealing for categorical attributes, considering each and every unique value of a numerical feature is cumbersome. %
Moreover, doing so may not reflect the underlying human thought process: as an example, consider ``high sugar content'' in contrast to ``70g of sugar per 100g of a product'', with both 0g and 100g in the latter case encoded as an \emph{absent} concept in the corresponding IR.%

\begin{figure}[t]%
    \centering
    \begin{subfigure}[t]{0.45\textwidth}
        \centering
        \includegraphics[width=.75\textwidth]{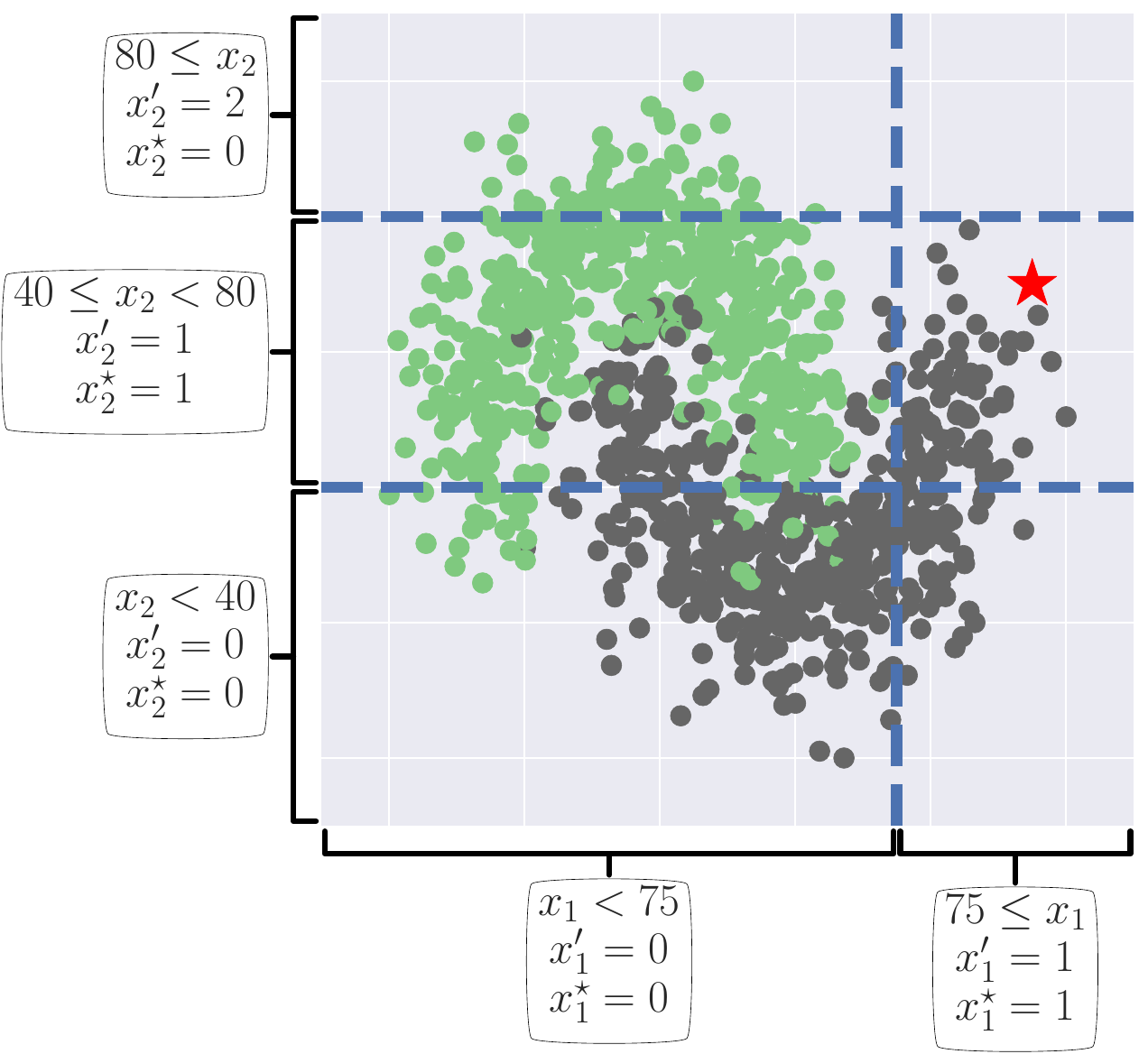}%
        \caption{Discretised and binarised numerical features become the components of the tabular interpretable representation (\(x^\star\)).\label{fig:tab_ex:1}}
    \end{subfigure}
    \hspace{0.033333333\linewidth}
    \begin{subfigure}[t]{0.45\textwidth}
        \centering
        \includegraphics[width=0.66666667\textwidth]{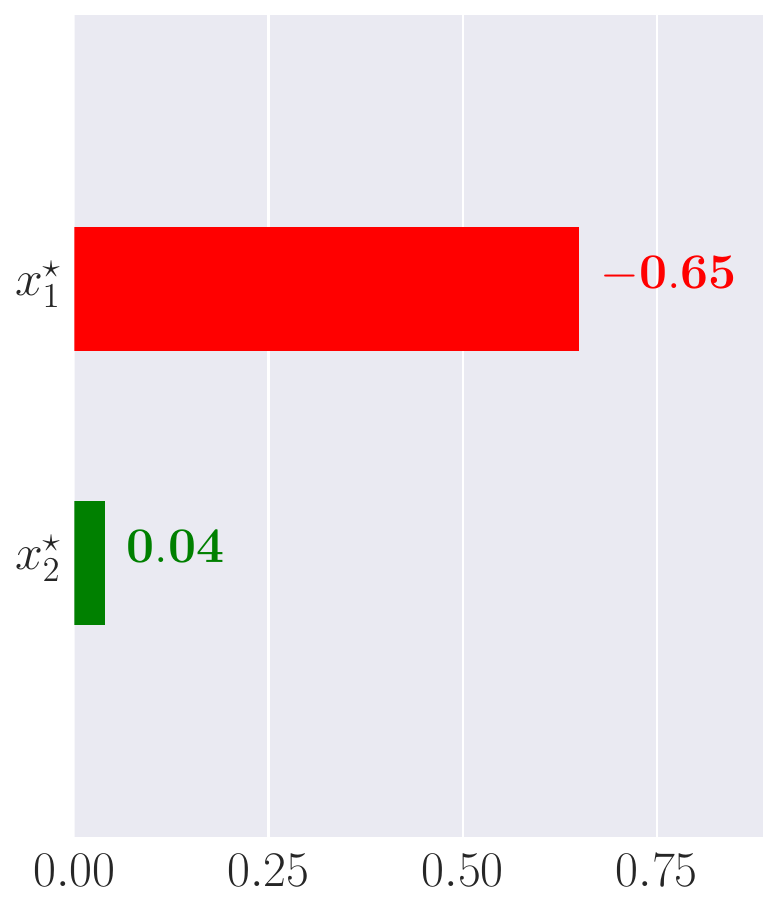}%
        \caption{Explanations show influence of IR components on predicting the \emph{grey} class for the red \(\star\) instance and, more generally, the entire \(x^\star = (1, 1)\) hyper-rectangle.\label{fig:tab_ex:2}}%
    \end{subfigure}
    \caption{%
Example of an influence-based explanation of tabular data with the interpretable representation built upon \emph{discretisation} (\(x^\prime\)) of numerical features followed by \emph{binarisation} (\(x^\star\)). %
Panel~(\subref{fig:tab_ex:1}) illustrates a synthetically generated \emph{two moons} toy data set with two numerical features \((x_1, x_2)\) and two classes denoted by grey and green dots. %
The red \(\star\) represents the explained instance \(x\), the dashed blue lines mark the attribute binning (discretisation), \(x^\prime\) is the (intermediate) discrete representation, and \(x^\star\) encodes the binary IR created for the \(\star\) data point. %
Panel~(\subref{fig:tab_ex:2}) depicts the magnitude of the influence that \(x_1^\star: 75 \leq x_1\) and \(x_2^\star: 40 \leq x_2 < 80\) have on the underlying black-box model predicting the \emph{grey} class for the \(\star\) instance (and more broadly any other data point residing within the same hyper-rectangle).%
\label{fig:tab_ex}}%
\end{figure}

Building on this observation, a natural extension of such a tabular interpretable representation is to discretise numerical features into (meaningful) categorical bins, e.g., \(x_2 < 40\), \(40 \leq x_2 < 80\) and \(80 \leq x_2\), %
which procedure establishes high-level intelligible concepts~\cite{garcia2012survey,kotsiantis2006discretization}. %
(This step is not needed for categorical attributes, which already have discrete values.) %
Since a binary representation allows to encode only two events for each attribute, we need to define the corresponding information removal proxy, which determines the meaning of a discrete concept being \emph{present} or \emph{absent}. %
Given that the IR should be specific to the explained data point, the binary on/off vector is constructed to indicate whether a feature value of an arbitrary instance \(x\) is of the \emph{same} (\(1\) for a \emph{present} concept) or \emph{different} (\(0\) for an \emph{absent} concept) category as the corresponding attribute of the data point selected to be explained \(\mathring{x}\). %
For example, if the second feature \(x_2\) of the explained instance \(\mathring{x}\) is \(\mathring{x}_2 = 70\), based on the aforementioned bin boundaries, any instance \(x\) whose second attribute is within the \(40 \leq x_2 < 80\) range is assigned \(x^\star_2 = 1\) in the binary IR, and \(0\) otherwise, i.e.,%
\[
x^\star_i =
\begin{cases}
    1    & \text{if~} x_i \text{~and~} \mathring{x}_i \text{~belong to the same numerical bin or have the same categorical value,} \\
    0    & \text{otherwise.}
\end{cases}
\]
This procedure is depicted in Figure~\ref{fig:tab}. %
When paired with a surrogate linear model, this interpretable representation allows us to investigate the influence of the designated concepts -- i.e., each numerical feature being within the specified range and each categorical attribute being of the particular value -- on the black-box prediction of the data point selected to be explained, or more precisely \emph{any} instance located within the same hyper-rectangle -- see Figure~\ref{fig:tab_ex} for an example. %

Notably, such a binary IR of tabular data is specific to the explained instance (more generally, its hyper-rectangle), as was the case for images and text. %
Nonetheless, the discretisation underlying tabular data can be easily reused for explaining other instances from the same data set. %
While a common practice~\cite{ribeiro2016why}, it should be added that such an approach might affect the faithfulness of the resulting insights since the goal is to produce a \emph{local} explanation of the selected data point, hence the discretisation should explicitly model the characteristics of the explained neighbourhood. %
Reusing the same discretisation to generate instance-specific IRs for tabular data can be compared to creating a super-pixel partition of a particular image and then reapplying it to other, \emph{unrelated} images, yielding a conceptually flawed interpretable representation. %
Additionally, selecting the bin boundaries when discretising tabular data, as well as grouping values of categorical attributes, is non-trivial and might bias the explanation similar to the influence of text pre-processing and tokenisation steps or image segmentation and occlusion strategies. %
Since neither globally nor locally faithful discretisation can capture uniqueness of a black-box decision boundary universally well for an arbitrary data subspace, each explained instance requires bespoke discretisation of the feature space~\cite{sokol2019blimey}. %

\section{Configuring Interpretable Representations\label{sec:ir-analysis}}%

Interpretable representations of text data that are based on token removal are appealing from an XAI perspective; they can be easily expanded with relevant natural language processing techniques without the need for any computational proxy, which makes these transformations largely consistent with human intuition in the explainability context. %
Image data, on the other hand, lack such a seamless operationalisation of IRs, forcing us to use an occlusion-based proxy to discard information from individual segments. %
This poses several challenges for the trustworthiness, robustness and computational soundness of the resulting explanations as well as their consistency with the explainees' intuition, especially since segmentation algorithms cannot rely on natural separation criteria such as whitespace characters and autonomy of words found in text data. %
In particular, we are faced with parameterising these IRs based on \emph{segmentation granularity} and \emph{occlusion strategy}, with certain choices possibly exhibiting undesired properties or being ineffective in ``removing'' super-pixels. %
The masking colour may impact the veracity of explanations, regardless of the underlying occlusion approach, since these insights rely on an implicit assumption that the black-box model is \emph{neutral} with respect to the occlusion colour, i.e., none of the modelled classes is biased towards it~\cite{mittelstadt2019explaining}. %
Adjusting the segmentation granularity can also play an important role given high correlation of adjacent super-pixels.%

In contrast, tabular data require by far the most complicated interpretable representation whose explanatory meaning may be difficult to grasp due to the counterintuitive process of ``switching off'' interpretable concepts. %
Moreover, the underlying information removal proxy requires discretisation of continuous features followed by a binarisation step -- a procedure that results in information loss and is sensitive to the selection of binning thresholds. %
Given its significance, the parameterisation of both image and tabular IRs should be explicitly optimised based on clearly defined objectives that appreciate the uniqueness of the problem at hand and recognise interpretable representations as independent entities and vital components of (surrogate) explainers~\cite{sokol2024what}. %
Out of these three IRs, the one for text has the advantage of allowing the tokens to be \emph{truly} removed from a sentence (although this is more of a property of the underlying predictive model rather than the interpretable representation itself). %
Specifically, text classifiers are more flexible and do not assume input of a fixed size, while vision models cannot handle missing pixels and tabular predictors usually require all features to be present. %
Investigating the effects of text pre-processing and tokenisation on the quality of the corresponding IRs is outside of the scope of this paper since it is a multifaceted endeavour, a narrow study of which may not provide comprehensive insights given the sheer quantity and diversity of available techniques.%

The interpretable representations of image and text data discussed here are \emph{implicitly local} -- they are intended (and possibly valid) only for the data point (sentence or image) for which they were created -- whereas the scope of the tabular IR is more ambiguous. %
Another property that the former two IRs have and the latter lacks is \emph{determinism} of the underlying representation change (within the scope of a single instance) as demonstrated by Figures~\ref{fig:txt}, \ref{fig:img} and \ref{fig:tab}. %
Transforming images and text between the two domains only requires memorising the structure or skeleton of the explained data point: adjacency of segments and their original pixel values for images (assuming that the segmentation and occlusion strategies are fixed), and order of tokens as well as their pre-processing for text. %
Tabular data IRs, however, lack this one-to-one correspondence between an instance and its interpretable representation -- due to the aforementioned information loss caused by the discretisation and binarisation steps -- which can only be restored by mapping each data point to its IR coordinates and storing this correspondence table for future retrieval~\cite{sokol2019blimey}. %
While this workaround is possible when starting with an instance in the original representation, it cannot overcome stochasticity when we are only given the IR encoding of a data point. %
Notably, this property helps to guarantee uniqueness of explanations, which is important for their stability, hence preserving explainees' trust~\cite{rudin2019stop,sokol2020explainability,sokol2020limetree}.%

\subsection{Occlusion-based Interpretable Representations of Images\label{sec:appendix:img}}%

\begin{figure}[b]%
    \centering
    \begin{subfigure}[t]{0.325\textwidth}
        \centering
        \includegraphics[width=\textwidth]{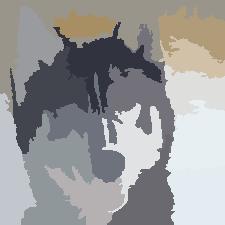}%
        \caption{17-segment partition.\label{fig:husky_mean_17}}
    \end{subfigure}
    \hfill
    \begin{subfigure}[t]{0.325\textwidth}
        \centering
        \includegraphics[width=\textwidth]{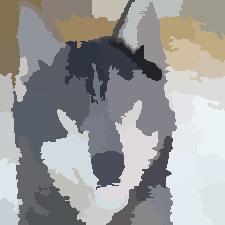}%
        \caption{51-segment partition.\label{fig:husky_mean_50}}
    \end{subfigure}
    \hfill
    \begin{subfigure}[t]{0.325\textwidth}
        \centering
        \includegraphics[width=\textwidth]{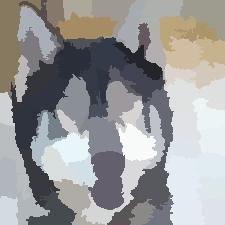}%
        \caption{71-segment partition.\label{fig:husky_mean_70}}
    \end{subfigure}
    \caption{%
\emph{Mosaic} or \emph{blurring} effect observed for the mean-colour occlusion strategy when ``removing'' all of the super-pixels with segmentation granularity increasing in size. %
The image was split into (\subref{fig:husky_mean_17}) 17, (\subref{fig:husky_mean_50}) 51 and (\subref{fig:husky_mean_70}) 71 super-pixels with the SLIC algorithm~\cite{achanta2012slic}, which performs \(k\)-means clustering in the colour space.%
\label{fig:husky_mean}}%
\end{figure}

Occlusion-based interpretable representations of images are parameterised by \emph{segmentation granularity} and \emph{colouring strategy}. %
The former allows the explanation to capture a desired level of detail, while the latter is used as a proxy for removing the content of super-pixels to hide the information they carry from a black box. %
The exact influence of these two properties on the resulting explanations therefore needs to be uncovered to inform the design of robust explainability techniques with well-understood behaviour. %
For example, consider the \textbf{mean-colour occlusion} used by the popular LIME explainer~\cite{ribeiro2016why}, which for some image partitions and super-pixel colour distributions may have undesired effects that undermine the utility of the occlusion procedure. %
With this approach, segments that have a relatively \emph{uniform colour gamut} may, effectively, be impossible to remove; this is especially common for fragments that are in the background or out of focus, e.g., bokeh and depth-of-field effects. %
\emph{Segmentation granularity} is also important: the smaller the segments are, the more likely it is that their colour composition is uniform given the ``continuity'' of images, i.e., high correlation of adjacent pixels. %
This \emph{mosaic} or \emph{blurring} effect is depicted in Figure~\ref{fig:husky_mean} for three different super-pixel granularity settings, showing how much of discriminative information is preserved in each case despite ``removing'' the content of \emph{all} the segments.%

\begin{figure}[t]%
    \centering
    \begin{subfigure}[t]{0.325\textwidth}
        \centering
        \includegraphics[width=\textwidth]{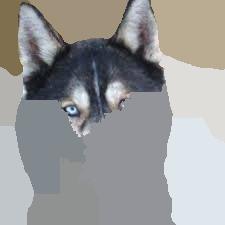}%
        \caption{Mean-colour occlusion.\label{fig:husky_patchwork_mean}}
    \end{subfigure}
    \hfill
    \begin{subfigure}[t]{0.325\textwidth}
        \centering
        \includegraphics[width=\textwidth]{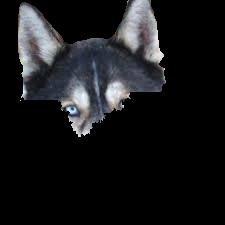}%
        \caption{Black occlusion.\label{fig:husky_patchwork_bk}}
    \end{subfigure}
    \hfill
    \begin{subfigure}[t]{0.325\textwidth}
        \centering
        \includegraphics[width=\textwidth]{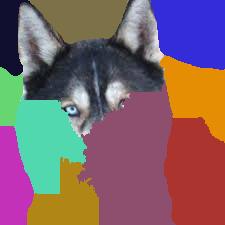}%
        \caption{Random-patch occlusion.\label{fig:husky_patchwork_rand}}%
    \end{subfigure}
    \caption{%
Image occlusion strategy influences the resulting explanations. %
The picture shown in Figure~\ref{fig:img_ex:1} is classified by a black box as \emph{Eskimo dog} with 84\% probability. %
Based on 11 super-pixels, the (\subref{fig:husky_patchwork_mean}) mean-colour occlusion of all the segments but one results in 78\%, (\subref{fig:husky_patchwork_bk}) black occlusion in 15\% and (\subref{fig:husky_patchwork_rand}) random-patch occlusion in 56\% probability of the same class. %
These results show that the mean occlusion strategy cannot effectively remove information from this particular image; the random-patch approach preserves segment edges, which are quite revealing in this case; and the black occlusion is relatively good at removing the content of super-pixels.%
\label{fig:husky_patchwork}}%
\end{figure}

Occluding each image fragment with a different colour manifests another issue, namely the \emph{preservation of super-pixel contours}. %
This effect can be observed in Figures~\ref{fig:husky_patchwork_mean} and \ref{fig:husky_patchwork_rand} respectively for the mean and random-patch occlusion strategies. %
Notably, whenever the segmentation coincides with objects' edges or regions of an image where colour continuity is not preserved -- which is common for edge-based segmenters -- replacing super-pixels with their mean or a random colour causes (slight) colour variations between adjacent segments. %
These artefacts emphasise edges in a (partially) occluded image that may at times convey enough information for a black-box model to correctly recognise its class; for example, see Figures~\ref{fig:husky_patchwork_mean} and \ref{fig:husky_patchwork_rand} where despite replacing all the super-pixels but \#4 with mean and random colours respectively, the model predicts the original class with 78\% and 56\% probability down from 84\% for the unaltered photo. %
Since most of these issues are consequences of using the random-patch or mean-colour occlusion, it may seem that fixing a single masking colour for all of the segments would eradicate some of these problems. %
Such an approach hides the edges between occluded super-pixels and removes their content instead of just ``blurring'' the image, which is the case for the mean-colouring strategy. %
However, the edges between occluded and preserved segments remain visible -- see Figure~\ref{fig:husky_patchwork_bk}, which depicts using black occlusion that yields only 15\% probability of the original class -- and choosing a neutral colour that does not bias the explanations -- e.g., the relation between blue and objects such as the sky or bodies of water -- remains an open question. %
Notably, the problem of selecting a \emph{reference point} or \emph{foil} for explanations is not unique to occlusion-based interpretable representations of images and it is particularly problematic for tabular data~\cite{mittelstadt2019explaining}.%

\begin{figure}%
    \centering
    \begin{subfigure}[t]{0.46\textwidth}
        \centering
        \includegraphics[width=\textwidth]{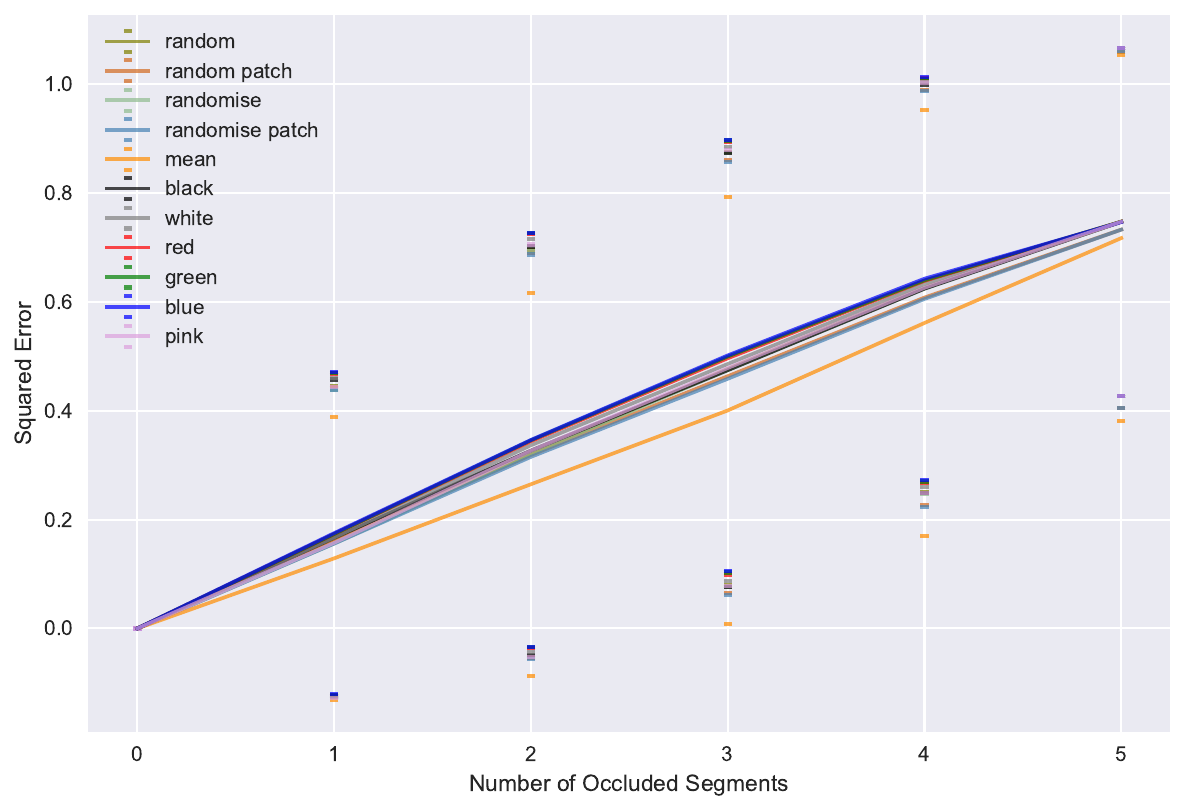}%
        \caption{5-segment partition run on 749 images.\label{fig:img_exp:5}}%
    \end{subfigure}
    \hfill
    \begin{subfigure}[t]{0.46\textwidth}
        \centering
        \includegraphics[width=\textwidth]{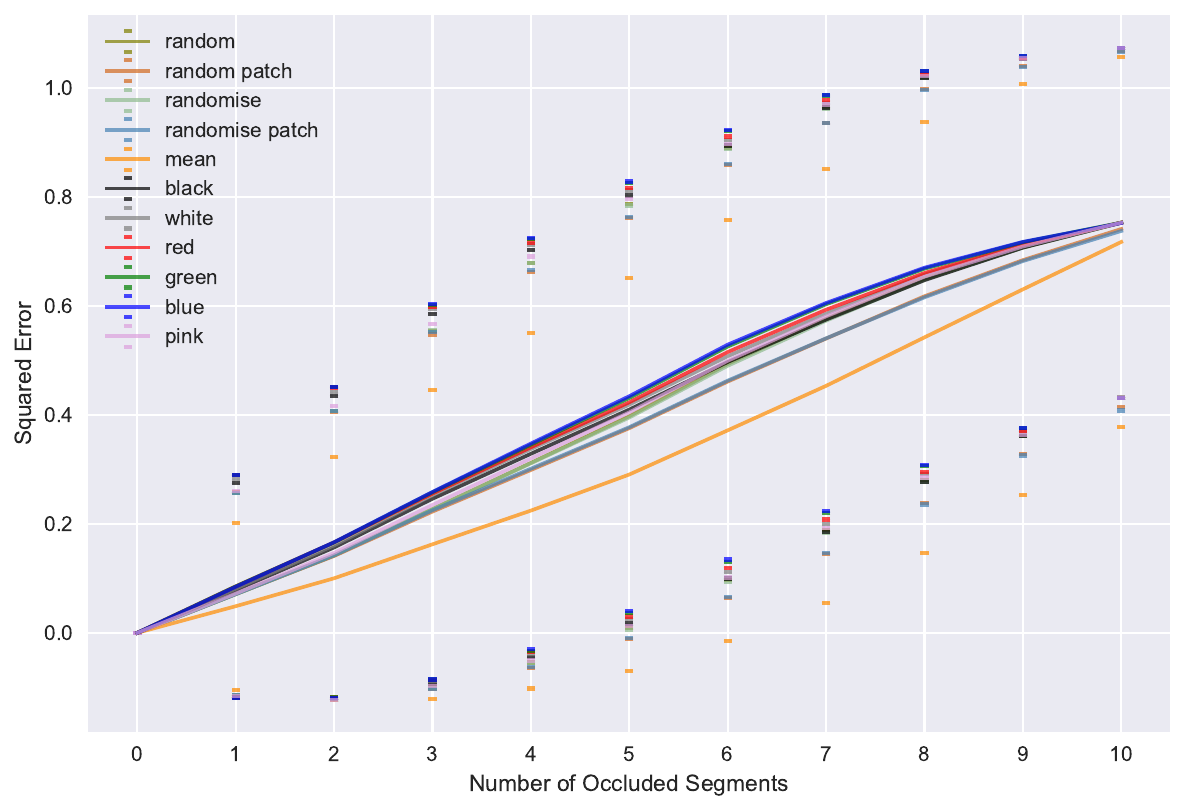}%
        \caption{10-segment partition run on 873 images.\label{fig:img_exp:10}}%
    \end{subfigure}
    \begin{subfigure}[t]{0.46\textwidth}
        \centering
        \includegraphics[width=\textwidth]{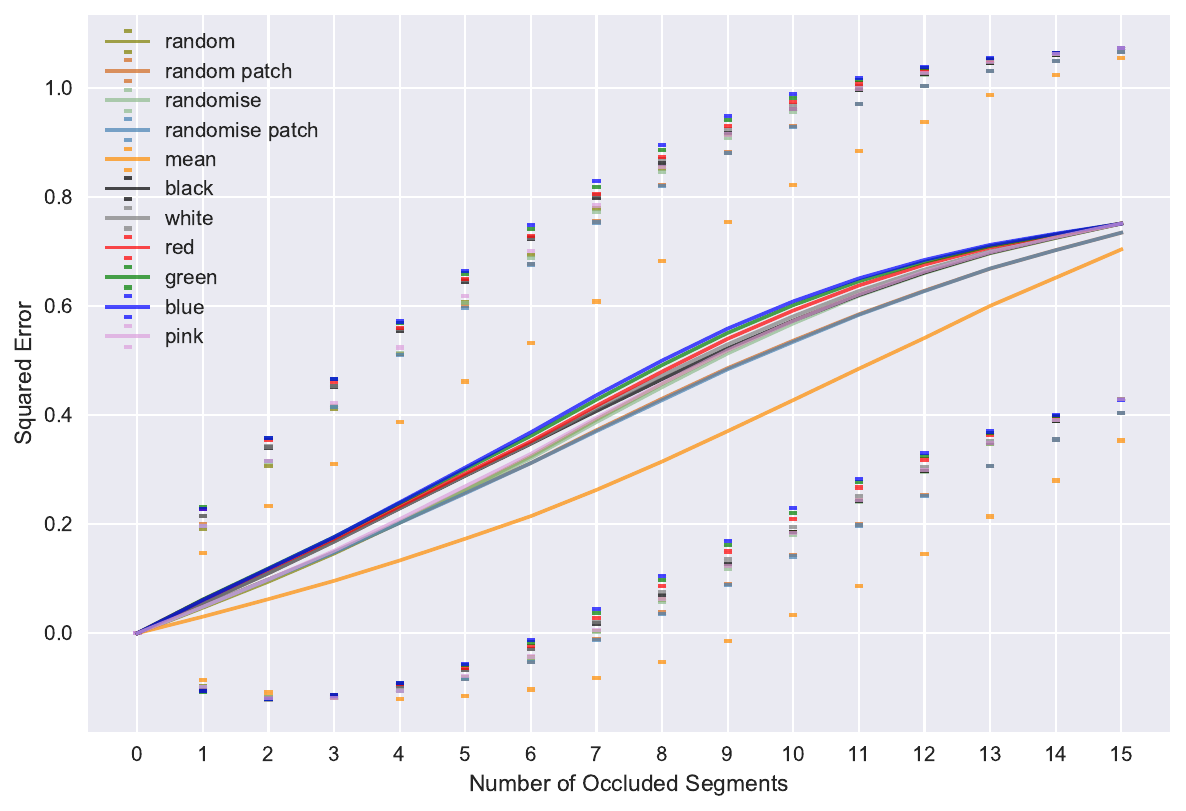}%
        \caption{15-segment partition run on 801 images.\label{fig:img_exp:15}}%
    \end{subfigure}
    \hfill
    \begin{subfigure}[t]{0.46\textwidth}
        \centering
        \includegraphics[width=\textwidth]{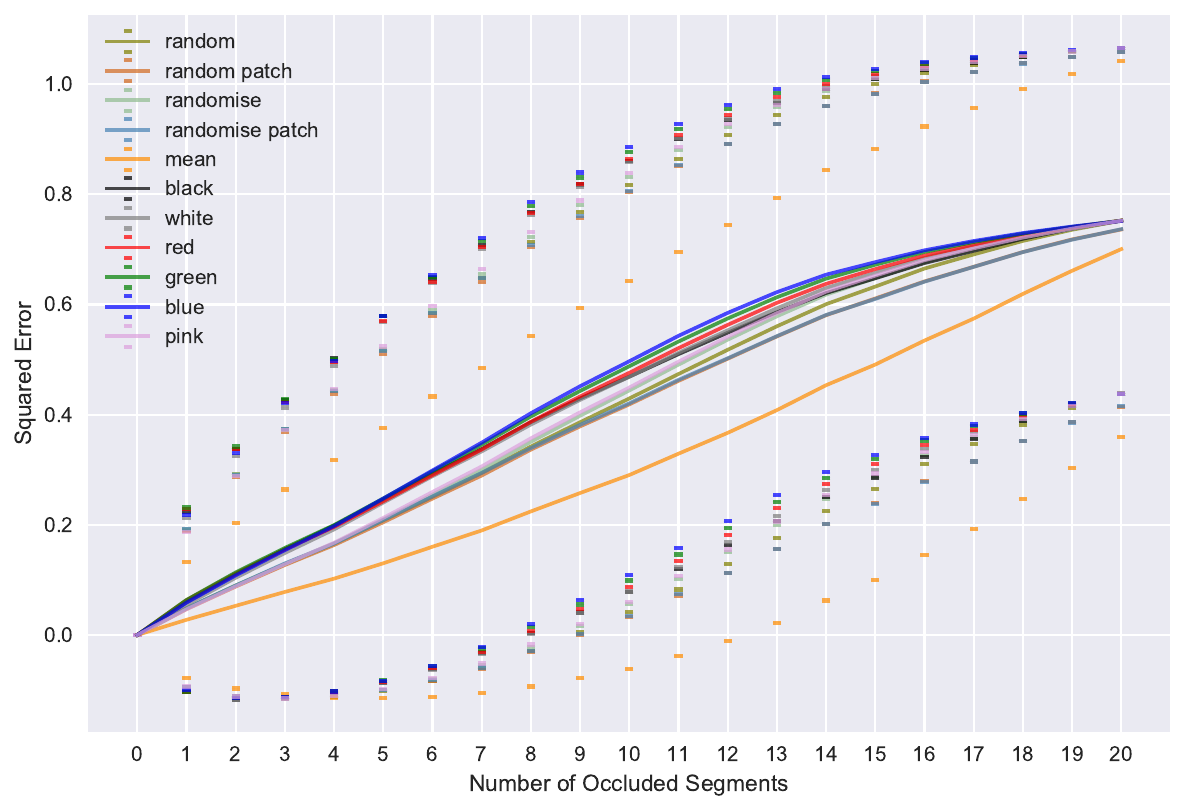}%
        \caption{20-segment partition run on 884 images.\label{fig:img_exp:20}}%
    \end{subfigure}
    \begin{subfigure}[t]{0.46\textwidth}
        \centering
        \includegraphics[width=\textwidth]{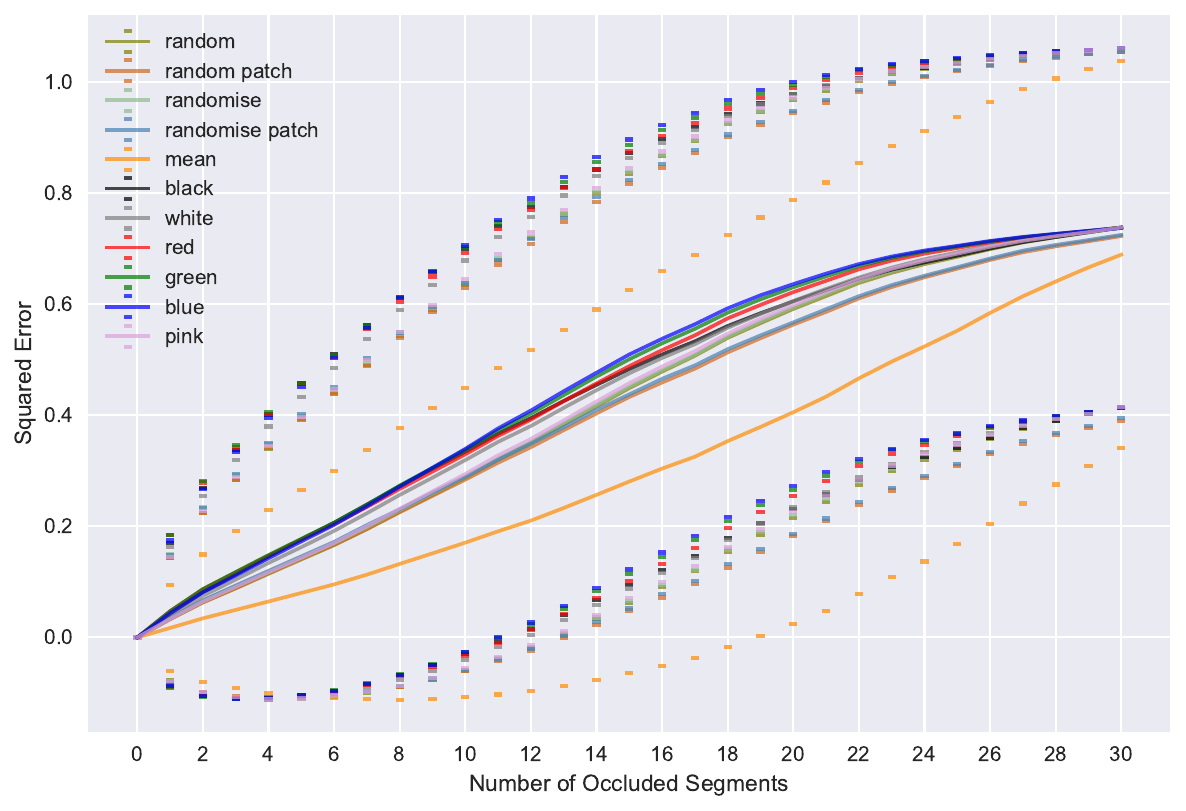}%
        \caption{30-segment partition run on 710 images.\label{fig:img_exp:30}}%
    \end{subfigure}
    \hfill
    \begin{subfigure}[t]{0.46\textwidth}
        \centering
        \includegraphics[width=\textwidth]{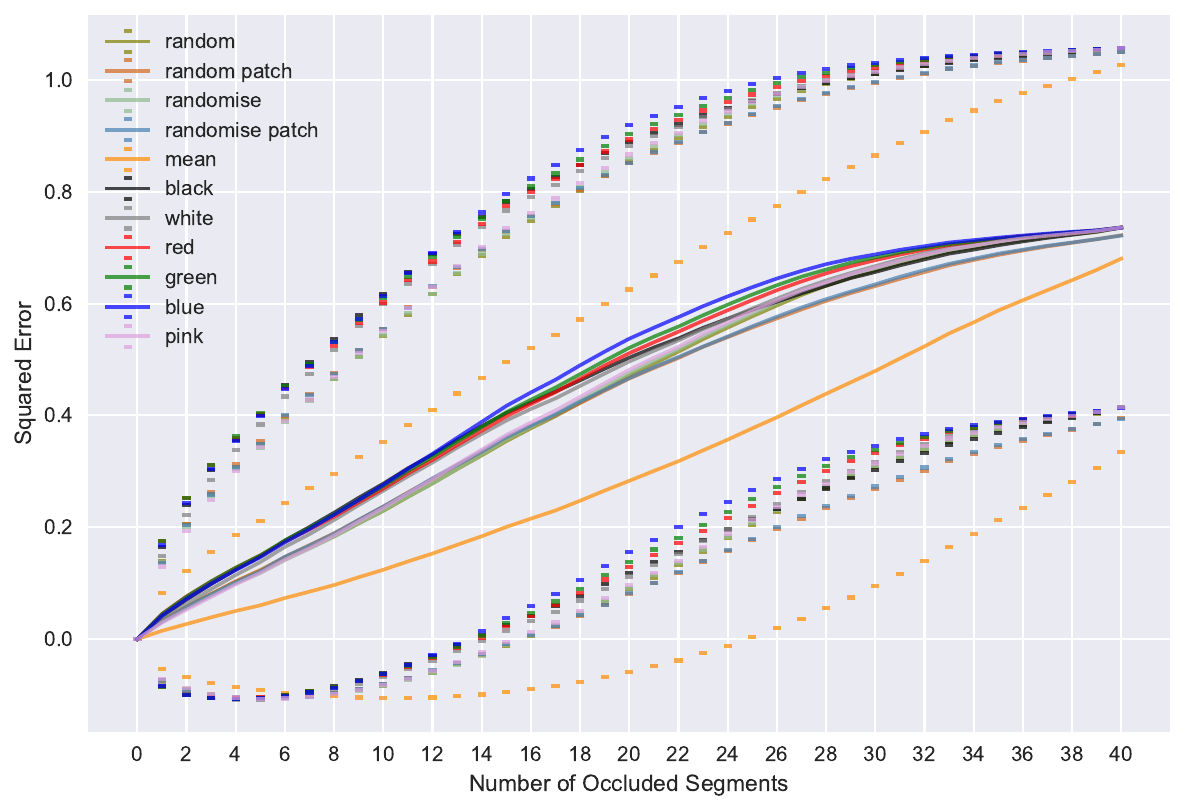}%
        \caption{40-segment partition run on 605 images.\label{fig:img_exp:40}}%
    \end{subfigure}
    \caption{%
Squared error (y-axis) calculated between the top prediction of an image (probability estimate) and predictions of the same class when incrementally occluding a higher number of random super-pixels (x-axis) with a given colouring strategy (legend). %
The segmentation is based on the SLIC algorithm~\cite{achanta2012slic}; the number of images used for each experiment is recorded in the captions above; a random sample of 100 occlusion patterns was generated for each step with a fixed number of super-pixel occlusions. %
The curves capture the mean of individual squared errors, with their standard deviation depicted by horizontal bars of the same colour -- a lower value indicates that the black box is better able to predict the top class despite information removal. %
The panels show that the mean occlusion strategy is not as effective at hiding information from the black box as using a single, random or randomised colour to the same end. %
The plots also reveal that when an image is split into more segments, the ineffectiveness of the mean-colouring approach gets magnified due to the increased colour uniformity of individual super-pixels (see Figure~\ref{fig:husky_mean} for an example of this phenomenon).%
\label{fig:img_exp}%
}
\end{figure}

\paragraph{Experiment Setup}%
To capture the influence of these characteristics -- \textbf{colour uniformity}, \textbf{segmentation size} and \textbf{edge visibility} -- on the effectiveness of the information removal proxy, we design and execute ablation studies. %
Empirically quantifying the effect of the colouring strategy and segmentation granularity on the ability of a black box to consistently predict a (partially distorted) image allows us to better understand the significance of these choices. %
To this end, we use images from the ImageNet~\cite{deng2009imagenet} validation set that are square and no smaller than 256\(\times\)256 pixels. %
Next, we resize them to 256\(\times\)256 pixels and segment them with the SLIC algorithm~\cite{achanta2012slic} -- which performs \(k\)-means clustering in the RGB (Red, Green, Blue) colour space -- using the implementation provided by scikit-image~\cite{walt2014scikit-image}. %
Since some of the images cannot be segmented into the desired number of super-pixels, only their subset (whose size is given in Figure~\ref{fig:img_exp}) is used for the study. %
For all of the experiments, our black box is the pre-trained \emph{Inception v3} neural network distributed with PyTorch~\cite{paszke2019pytorch}. %
The study is implemented using the FAT Forensics Python package~\cite{sokol2019fat,sokol2020fat}, with the experiment code published on GitHub\footnote{\url{https://github.com/So-Cool/bLIMEy/tree/master/DAMI_2024}}.%

Specifically, segment occlusion is done with the following selection of colouring strategies denoted in the RGB space:%
\begin{description}[labelindent=.5em,labelwidth=9.0em,labelsep=0pt,leftmargin=9.5em,font=\bfseries]%
    \item [black] \RGB{0}{0}{0};%
    \item [white] \RGB{255}{255}{255};%
    \item [red] \RGB{255}{0}{0};%
    \item [green] \RGB{0}{255}{0};%
    \item [blue] \RGB{0}{0}{255};%
    \item [pink] \RGB{255}{192}{203};%
    \item [mean] each super-pixel is replaced with a solid patch of the mean colour computed for the pixels residing within this segment;%
    \item [random] a \emph{single} random colour, sampled uniformly from the RGB space, is used to occlude all super-pixels across all experiments for a single image and fixed segmentation size;%
    \item [random patch] a \emph{separate} random colour, sampled uniformly from the RGB space, is used to occlude each individual super-pixel across all experiments for a single image and fixed segmentation size;%
    \item [randomise] a \emph{single} random colour, sampled uniformly from the RGB space, is generated for each individual occlusion pattern; and%
    \item [randomise patch] a \emph{separate} random colour, sampled uniformly from the RGB space, is generated for each individual super-pixel.%
\end{description}
The test images are partitioned into 5, 10, 15, 20, 30 and 40 regions to capture the influence of the segmentation granularity on the IR -- these tiers are visualised in separate panels of Figure~\ref{fig:img_exp}. %
For a fixed number of segments, we iterate over the quantity of occluded super-pixels from 0 to all of the partitions (x-axes in Figure~\ref{fig:img_exp}), randomising the occlusion pattern 100 times at each step. %
We apply this procedure to all of our test images, separately for every colouring strategy. %
Finally, we measure the influence of each occlusion strategy and segmentation granularity by calculating the \emph{squared error} -- \(\mathit{SE} = (y_i - \hat{y}_i)^2\) -- between the probability of the top class predicted for the unaltered image and the prediction of the same class when the image is (partially) occluded. %
We aggregate these scores by computing their mean and standard deviation (y-axes in Figure~\ref{fig:img_exp}), a low value of which indicates that the model can still predict the data point despite a distortion.%

\paragraph{Occlusion Colour}%
Figure~\ref{fig:img_exp} provides clear evidence that the \emph{mean} occlusion strategy behaves unlike any other approach, including all of the methods based on \emph{random} colour selection; additionally, there is no perceptible difference between the remaining colouring strategies -- their squared error curves are bundled together. %
More precisely, the lower metric value for the \emph{mean} technique indicates that it is not as effective at removing class-identifying information from images as any other occlusion strategy that we tested. %
Intuitively, the reason for this behaviour is the aforementioned \emph{blurring} or \emph{mosaic} effect depicted in Figure~\ref{fig:husky_mean}. %
This phenomenon becomes especially pronounced when images are segmented into smaller super-pixels, as having more of them for a fixed image size makes each partition more uniform with respect to the colour of its individual pixels -- the increasing separation of the squared error curve for the \emph{mean} strategy when moving from 5 (Figure~\ref{fig:img_exp:5}) to 40 (Figure~\ref{fig:img_exp:40}) segments. %
Additionally, Figure~\ref{fig:img_exp} illustrates the consequences of preserving contour lines between segments when occluding them with patches of different colour -- an example of which is visualised in Figure~\ref{fig:husky_patchwork_rand}. %
This behaviour is captured by the \emph{random patch} and \emph{randomise patch} strategies, both of which exhibit a lower squared error than any other technique based on a single, possibly random, occlusion colour; nonetheless, this effect appears negligible across our experiments.%

\paragraph{Segmentation Granularity}%
By inspecting each panel of Figure~\ref{fig:img_exp}, we can see that the granularity of segmentation directly affects the \emph{mean}-colour occlusion strategy -- the aforementioned separation between the squared error curve of the \emph{mean} approach and every other curve. %
The behaviour of all the fixed-colour approaches, on the other hand, is very similar for any number of segments regardless of the exact occlusion colour (including its random selection) -- these squared error curves are clustered together in Figure~\ref{fig:img_exp}. %
Notably, this observation also applies to the \emph{random-patch} and \emph{randomise-patch} strategies, which methods reveal segment boundaries and can be very volatile given their random assignment of the occlusion colour to each individual super-pixel. %
Both of these insights offer clear evidence that using the \emph{mean} colouring should be avoided in occlusion-based interpretable representations of images. %
Figure~\ref{fig:img_exp} substantiates our observation that this occlusion strategy becomes less effective as the number of super-pixels increases since relatively small segments tend to have a uniform colour distribution because of the pixel \emph{continuity} -- i.e., high correlation of neighbouring pixels -- making them visually similar to their respective \emph{mean}-coloured patches. %
Additionally, this undesired phenomenon may affect images that have an out-of-focus background, e.g., portraits, since their blurry regions will be difficult to remove with the \emph{mean}-colour occlusion strategy.%

\paragraph{Snow in the Background}%
Observing the influence of each algorithmic component on the effectiveness of occlusion-based interpretable representations for images, hence explainers built upon them such as LIME, has prompted us to re-examine some of the conclusions drawn by \citet[\S6.4]{ribeiro2016why}. %
In particular, the inability of the mean occlusion strategy to discard information -- especially so for uniform colour patches and high segmentation granularity -- casts doubts on the veracity of explanations generated in the famous study of snow (visible in the background of a picture) biasing predictions of a model deciding between a wolf and an Eskimo dog. %
The mosaic effect resulting from this removal proxy -- captured by Figure~\ref{fig:husky_mean} -- and the overall ineffectiveness of this approach -- exemplified by Figure~\ref{fig:husky_patchwork} -- demonstrate acute problems of the downstream explanations in such a classification scenario. %
Specifically, consider segments of this image showing snow, which are replaced with their respective mean colours, thus producing off-white patches that still resembles snow; for example, compare the bottom-left and the bottom-right super-pixels in Figures~\ref{fig:img_ex:1} and \ref{fig:husky_patchwork_mean}. %
These almost visually indistinguishable alterations are likely to prevent the explainer from capturing the change in the probability predicted by the model under investigation, as shown by our experiments and exemplified in Figure~\ref{fig:husky_patchwork}, affecting soundness of the resulting LIME explanations. %
While such techniques may generate insights into black-box classifiers and help us to uncover spurious correlations, if not tuned to the problem at hand, stock explainers may cause more harm than good~\cite{rudin2019stop}.%

\subsection{Discretisation-based Interpretable Representations of Tabular Data\label{sec:disc:tab}}%

\begin{figure}[t]%
    \centering
    \begin{subfigure}[t]{0.49\textwidth}  %
        \centering
        \includegraphics[width=\textwidth]{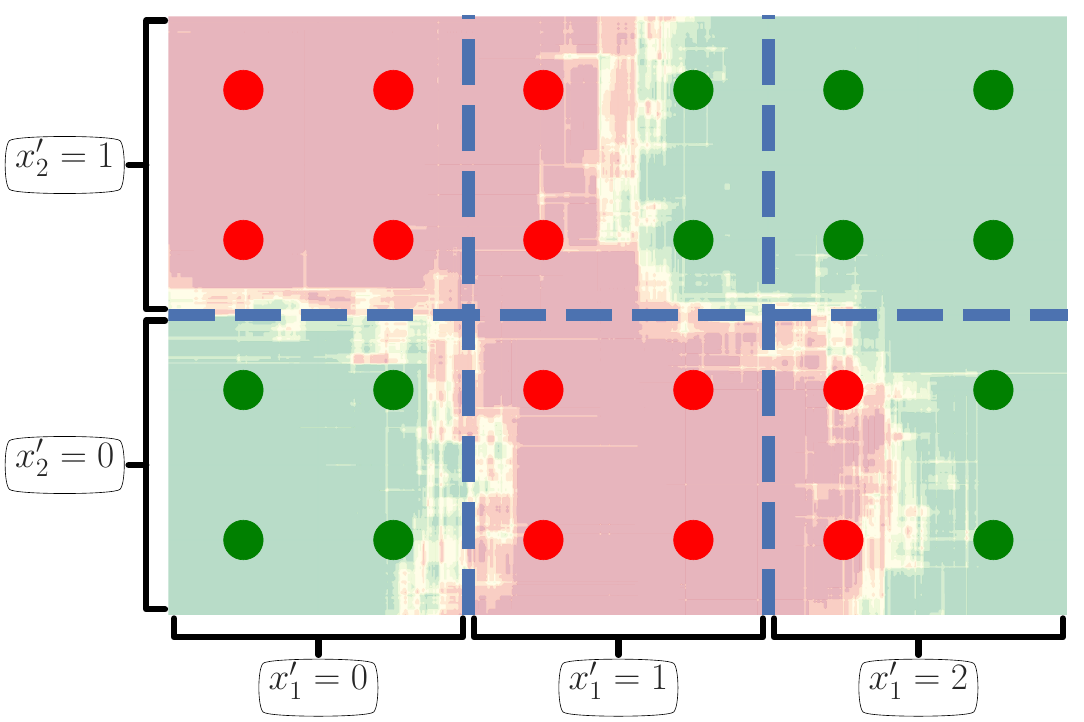}%
        \caption{\emph{Distribution}-aware discretisation.\label{fig:tabular_ir:dist}}%
    \end{subfigure}
    \hfill
    \begin{subfigure}[t]{0.49\textwidth}  %
        \centering
        \includegraphics[width=\textwidth]{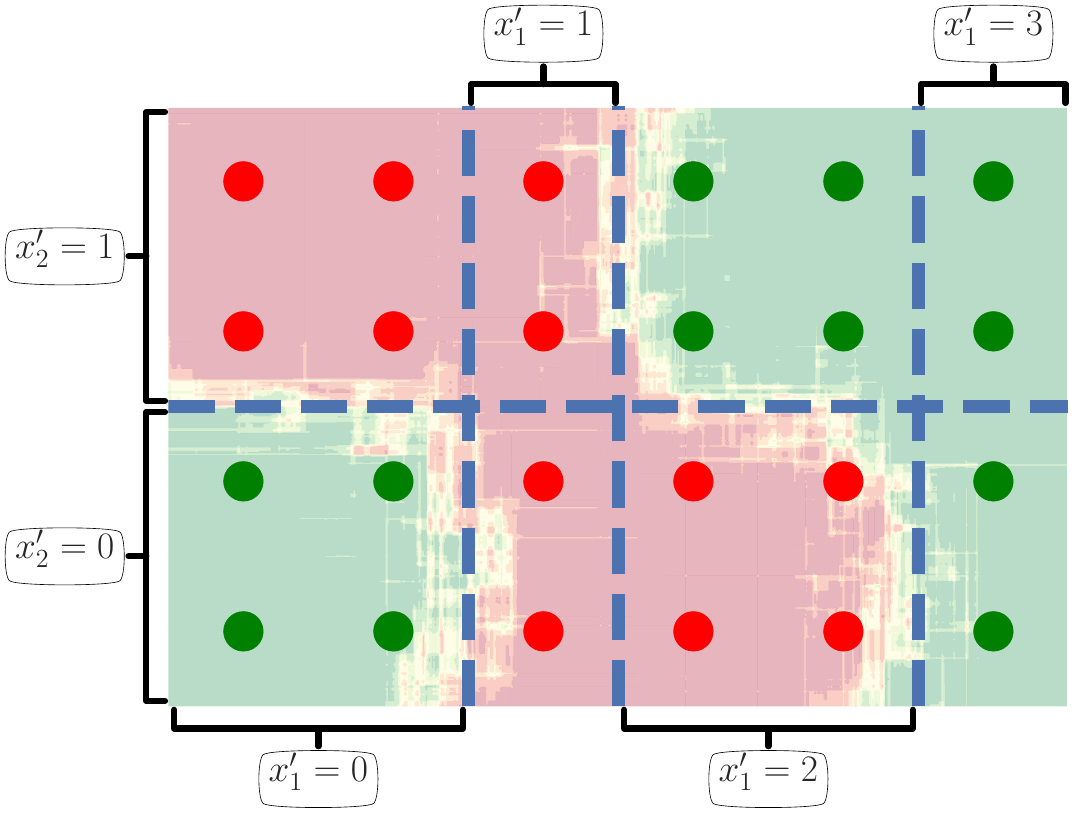}%
        \caption{\emph{Class}-aware (predictions) discretisation.\label{fig:tabular_ir:class}}%
    \end{subfigure}
    \caption{%
Discretisation is the main building block of interpretable representations for tabular data with numerical attributes. %
It can either be learnt based on data features alone -- an \emph{unsupervised} approach shown in Panel~(\subref{fig:tabular_ir:dist}) -- or additionally consider their black-box predictions -- a \emph{supervised} approach shown in Panel~(\subref{fig:tabular_ir:class}). %
These examples use a synthetic toy data set with two numerical features and 24 evenly-spaced instances whose class -- red or green -- is given by each point's colour; the background shading indicates the class prediction provided by the underlying (black-box) model; \(x^\prime_1\) and \(x^\prime_2\) encode the bin assignment for the first and second feature respectively. %
\label{fig:tabular_ir}}%
\end{figure}

The two factors that influence the creation of a binary interpretable representation of tabular data are the \emph{instance} selected to be explained -- which determines the reference hyper-rectangle -- and the ability of the \emph{discretisation} algorithm to (locally) approximate the black-box decision boundary -- which dictates its faithfulness~\cite{sokol2020explainability}. %
Only the latter property, however, is controlled algorithmically and can either be explicitly \emph{global}, i.e., learnt with respect to a whole data set, or \emph{local}, thus focusing on a specific neighbourhood. %
Furthermore, each variant can either observe just the \emph{data distribution}, or additionally take into account their \emph{black-box predictions}, presenting us with two distinct discretisation approaches:%
\begin{description}[labelindent=.5em,leftmargin=\parindent+.5em,font=\bfseries]%
    \item[distribution-aware (unsupervised)] based on the density of data in the local or global region chosen to be explained, e.g., quantile discretisation (Figure~\ref{fig:tabular_ir:dist}); and%
    \item[class-aware (supervised)] partitioning data according to a black-box decision boundary confined within the local or global region chosen to be explained (Figure~\ref{fig:tabular_ir:class}).%
\end{description}
While the scope and supervision level of a discretisation are the two main properties that affect the quality of a tabular interpretable representation, other aspects of this process can be considered as well, a summary of which can be found in relevant surveys~\cite{kotsiantis2006discretization,garcia2012survey}.%

\begin{figure}[t]%
    \centering
            \includegraphics[width=.3375\textwidth]{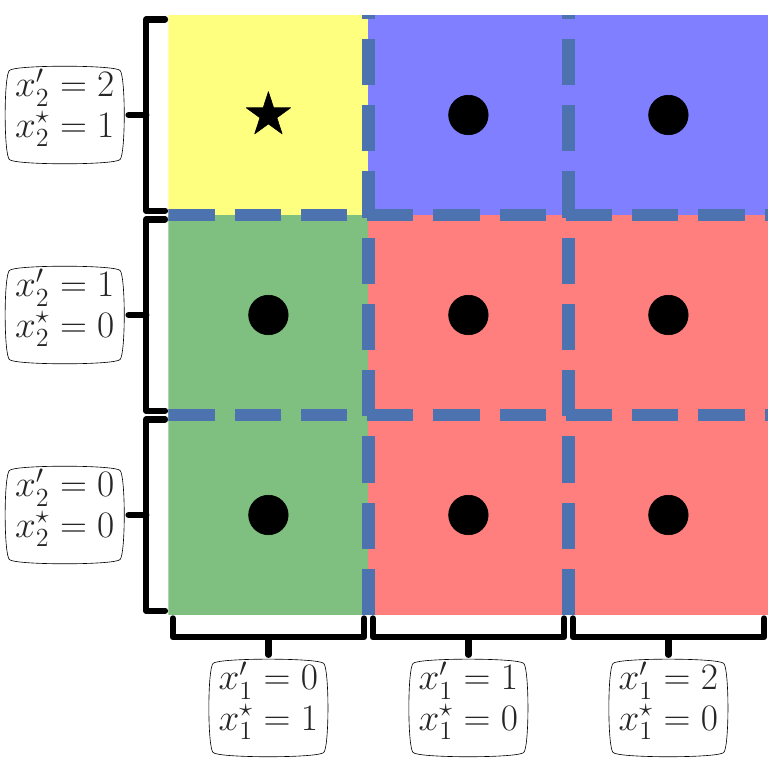}%
        \caption{%
Some hyper-rectangles \(x^\prime = (x_1^\prime, x_2^\prime)\) created through discretisation become indistinguishable in the binary interpretable representation \(x^\star = (x_1^\star, x_2^\star)\) of tabular data. %
The \(\star\) symbol indicates the explained instance and the background shading marks unique binary encodings \(x^\star = (x_1^\star, x_2^\star) \in \left\{(0, 0),\; (0, 1),\; (1, 0),\; (1, 1)\right\}\). %
This example uses a synthetic toy data set with two numerical features and nine evenly-spaced instances, one per hyper-rectangle created by the discretisation step. %
\label{fig:tabular_binary_ir}}%
\end{figure}

\paragraph{Information Loss}%
Discretisation and binarisation procedures tend to be many-to-one mappings. %
The intermediate \emph{discrete} representation of a tabular IR uniquely encodes each created hyper-rectangle -- see the \((x_1^\prime, x_2^\prime)\) coordinates in Figure~\ref{fig:tabular_binary_ir} -- \emph{explicitly} trading off precision for sparsity and intelligibility. %
However, the ensuing \emph{binarisation} step \emph{implicitly} loses information whenever a categorical feature has more than two unique values or a numerical attribute is partitioned into more than two intervals as shown by the background shading and the \((x_1^\star, x_2^\star)\) coordinates in Figure~\ref{fig:tabular_binary_ir}. %
Recall that for each of these binary IR features, \(1\) is assigned to the partition that contains the explained data point and \(0\) to all the other intervals, effectively making the latter categories \emph{indistinguishable}.%

The impossibility to discern data points belonging to different hy\-per-rec\-tan\-gles in the binary interpretable representation is particularly detrimental to the IR's ability to capture the intricacy of the black-box decision boundary. %
While the underlying discretisation may have closely approximated its shape, these details can be lost when transitioning into the binary space, especially if the decision boundary runs across hyper-rectangles that are merged in this process. %
For example, consider the discretisation shown in Figure~\ref{fig:tabular_ir:class}, assuming that the explained instance resides in the \(x^\prime = (1, 1)\) hyper-rectangle -- top row, second column from the left. %
In the binary representation, the remaining top-row hyper-rectangles \((0, 1)\), \((2, 1)\) and \((3, 1)\) would be bundled together -- akin to the process depicted by the background shading in Figure~\ref{fig:tabular_binary_ir} -- thus forfeiting the information that the first one belongs to the red class and the latter two to the green class. %
A similar grouping will happen in the bottom row, where \((0, 0)\), \((2, 0)\) and \((3, 0)\) will be merged. %
Observing this redundancy, nonetheless, can help us in search of a better mechanism to build tabular interpretable representations. %

\paragraph{Faithfulness}%
Since the predominant role of \emph{local surrogate explainers} is to approximate and simplify the behaviour of a black box near a selected instance, \emph{local} and \emph{class-aware} discretisation should be preferred. %
This procedure is a stepping stone towards representing interpretable concepts that are coherent with predictions of the underlying model, thus producing faithful and appealing insights. %
However, to the best of our knowledge, class-aware (supervised) discretisation approaches are absent in the explainability literature. %
Computationally, their objective can be expressed as \emph{maximising the purity or uniformity} of each hyper-rectangle with respect to the black-box predictions of data that it encloses -- this applies to regression as well as probabilistic and crisp classification models~\cite{kotsiantis2006discretization,garcia2012survey}. %
In particular, if the underlying task is crisp classification, we can use the \emph{Gini impurity} (\(\mathcal{L}_{\textnormal{G}}\)) defined in Equation~\ref{eq:appendix:tree:gini}, where \(H_i\) is a set of data points and their labels \((x, y)\) residing within the \(i^{\textnormal{th}}\) hyper-rectangle and \(C\) is the set of all the unique labels \(c\).%
\begin{gather}\label{eq:appendix:tree:gini}
  \begin{aligned}
    \mathcal{L}_{\textnormal{G}}(H_i) &= \sum_{c \in C} p_{H_i}(c) \times \left(1 - p_{H_i}(c)\right)\\
    p_{H_i}(c) &= \frac{1}{|H_i|} \sum_{(x, y) \in H_i} \mathds{1}_{y = c}%
  \end{aligned}
\end{gather}
On the other hand, when the task is regression or probabilistic classification (the formula applies separately to each individual class in the latter case), we can use the \emph{Mean Squared Error} (\(\mathcal{L}_{\textnormal{MSE}}\)) -- defined in Equation~\ref{eq:appendix:tree:mse} -- to quantify numerical uniformity of black-box predictions in each hyper-rectangle.%
\begin{gather}\label{eq:appendix:tree:mse}
  \begin{aligned}
    \mathcal{L}_{\textnormal{MSE}}(H_i) &= \frac{1}{|H_i|} \sum_{(x, y) \in H_i} (y - \widebar{y}_{H_i})^2\\%
    \widebar{y}_{H_i} &= \frac{1}{|H_i|} \sum_{(x, y) \in H_i} y%
  \end{aligned}
\end{gather}
When combining scores of multiple hyper-rectangles to assess the overall quality \(\mathcal{Q}\) of an interpretable representation, we opt for a weighted average of individual scores \(\mathcal{L}\) to account for the (possibly unbalanced) distribution of data points across these segments -- see Equation~\ref{eq:appendix:tree:score}.%
\begin{gather}\label{eq:appendix:tree:score}
  \mathcal{Q} = \frac{1}{\sum_{H_i} |H_i|} \sum_{H_i} |H_i| \times \mathcal{L}(H_i)%
\end{gather}
We use this formulation to evaluate the faithfulness of a quartile-based tabular IR (distribution-aware discretisation) used by the popular LIME explainer~\cite{ribeiro2016why}, and compare it with a simple tree-based IR (class-aware discretisation), testing both approaches in a global and local variant.%

\paragraph{Quartile IR}%
This interpretable representation is based on quartile discretisation of continuous features. %
The partition of the data space can either be global or local -- i.e., with respect the entire data set or its subset -- nonetheless each individual instance receives a distinct IR due to the binarisation step that follows. %
For each data point, global IRs are derived based on a shared discretisation computed for the entire data set. %
Local IRs, on the other hand, are composed separately for each instance in the data set based on samples located in its neighbourhood, which are used for the discretisation step. %
We rely on the formula given by Equation~\ref{eq:appendix:tree:score} to evaluate the faithfulness of both steps: discretisation and binarisation. %
For global discretisation this validation is performed on the entire data set. %
All the other approaches are assessed on a subset of data that, centred around the explained data point, is within the radius of 30 per cent of the maximum Euclidean distance computed between any two instances in the data set, which simulates locality of the explanation.%

\paragraph{Tree-based IR}%
This interpretable representation is based on a partition of the feature space learnt by a tree model. %
Its candidature stems from observing similarity between the tree learning objective and the proposed faithfulness evaluation metrics. %
Global analysis is performed by computing purity of the hyper-rectangles created by a tree fitted to the entire data set and validated on this training data as well as a local sample generated separately for each instance in the data set, which is a fair comparison given that the quartile-based IR can also access the whole data set. %
The local IR faithfulness, on the other hand, is calculated independently for each instance in the data set by learning a tree model on a subset of data that, centred around the explained instance, is within the radius of 30 per cent of the maximum Euclidean distance computed between any two instances in the data set, with the same data subset used to evaluate the quality of the resulting hyper-rectangles.%

\begin{figure}%
    \centering
    \begin{subfigure}[t]{0.95\textwidth}
        \centering
        \includegraphics[width=.623\textwidth]{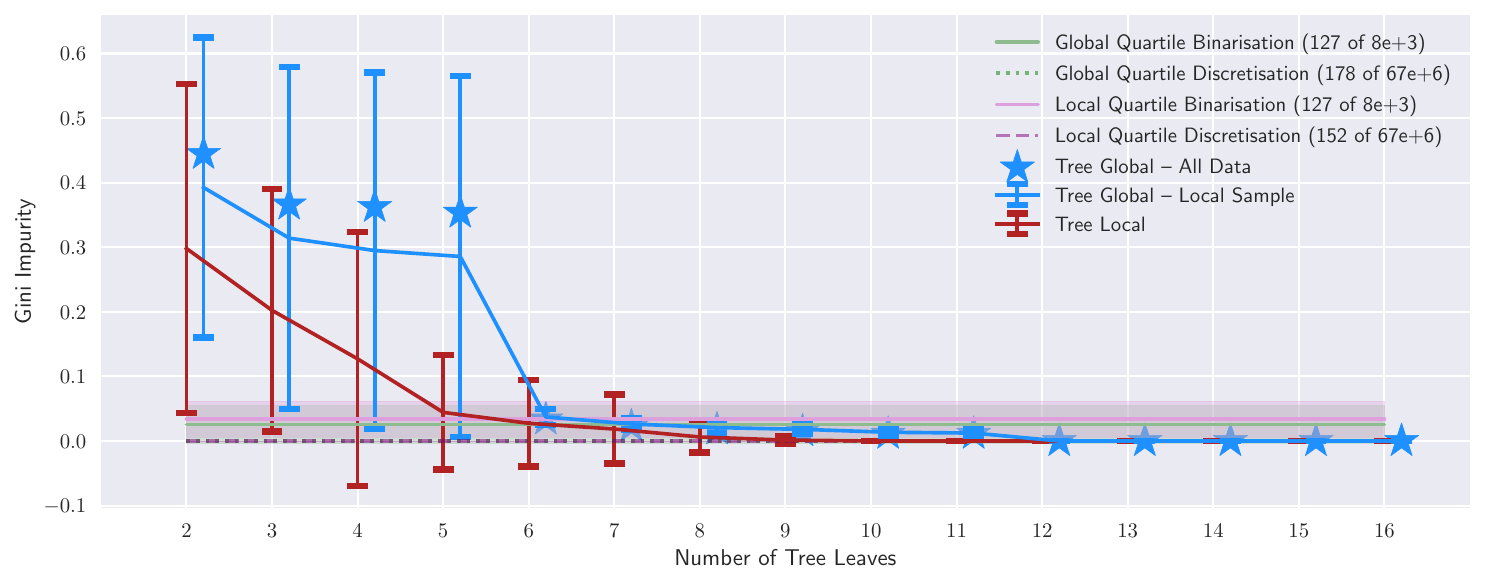}%
        \caption{Weighted average of the \emph{Gini impurity} computed for the \emph{wine} data IRs.\label{fig:tab_exp:wine}}%
    \end{subfigure}
    \begin{subfigure}[t]{0.95\textwidth}
        \centering
        \includegraphics[width=.623\textwidth]{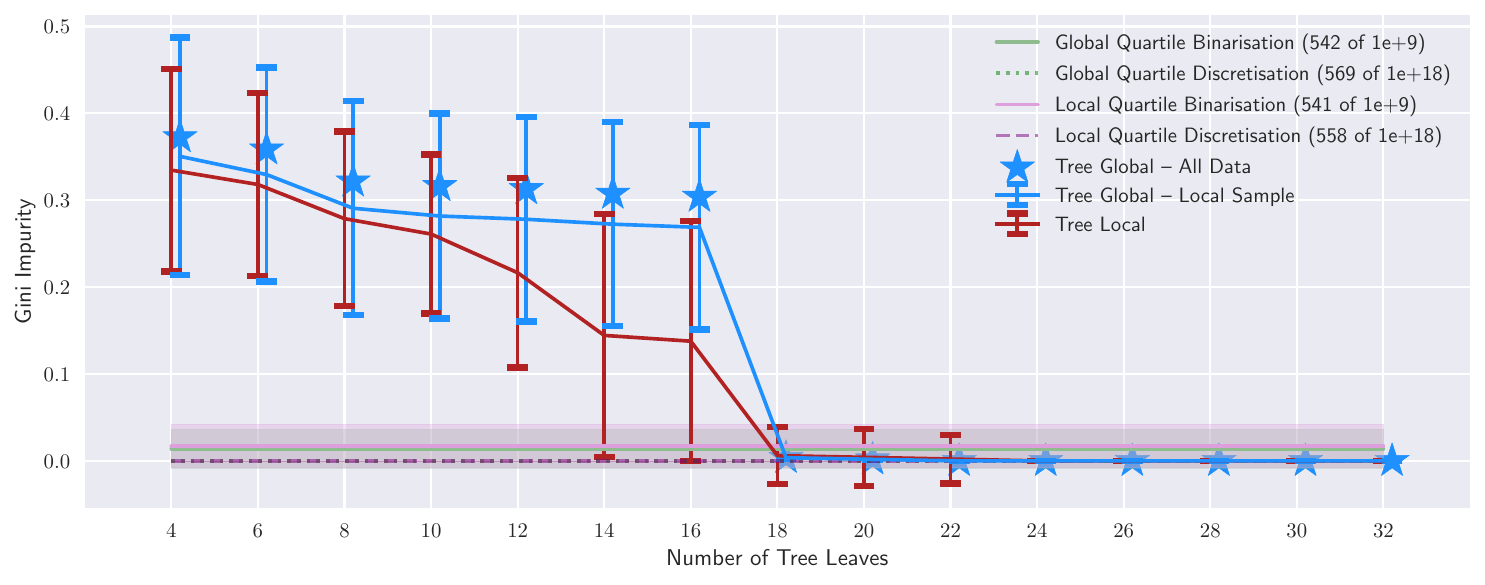}%
        \caption{Weighted average of the \emph{Gini impurity} computed for the \emph{cancer} data IRs.\label{fig:tab_exp:cancer}}%
    \end{subfigure}
    \begin{subfigure}[t]{0.95\textwidth}
        \centering
        \includegraphics[width=.623\textwidth]{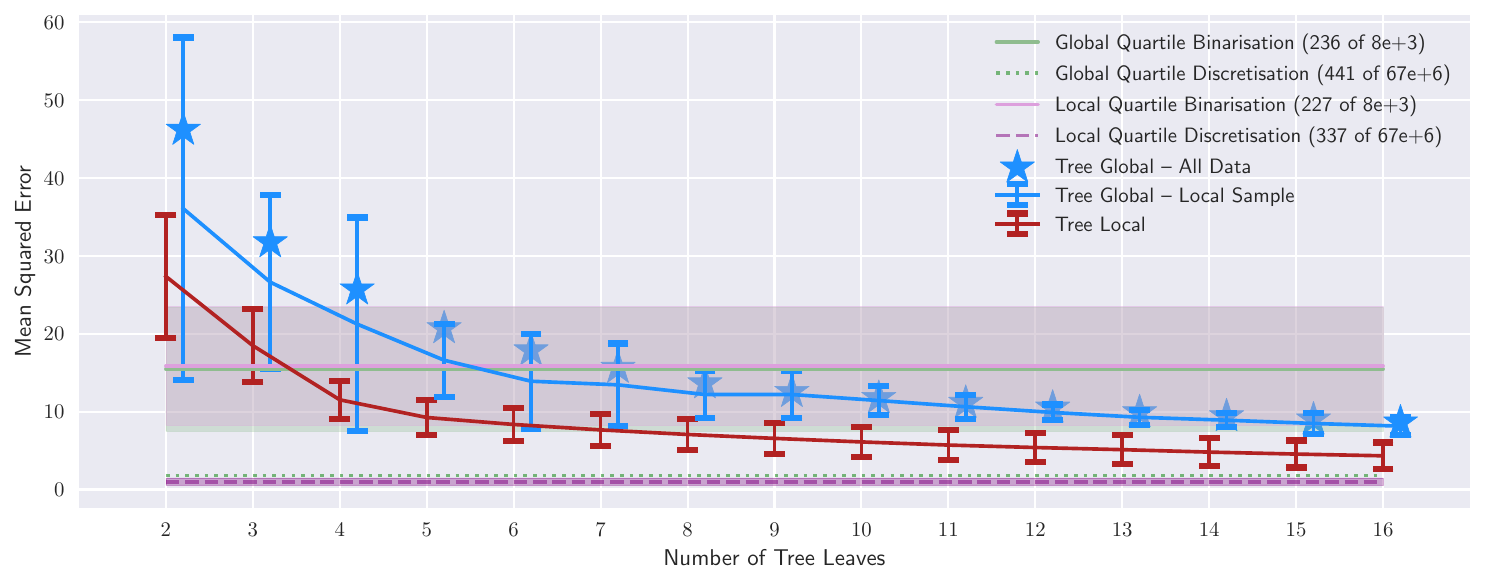}%
        \caption{Weighted average of the \emph{mean squared error} computed for the \emph{housing} data IRs.\label{fig:tab_exp:boston}}%
    \end{subfigure}
    \begin{subfigure}[t]{0.95\textwidth}
        \centering
        \includegraphics[width=.623\textwidth]{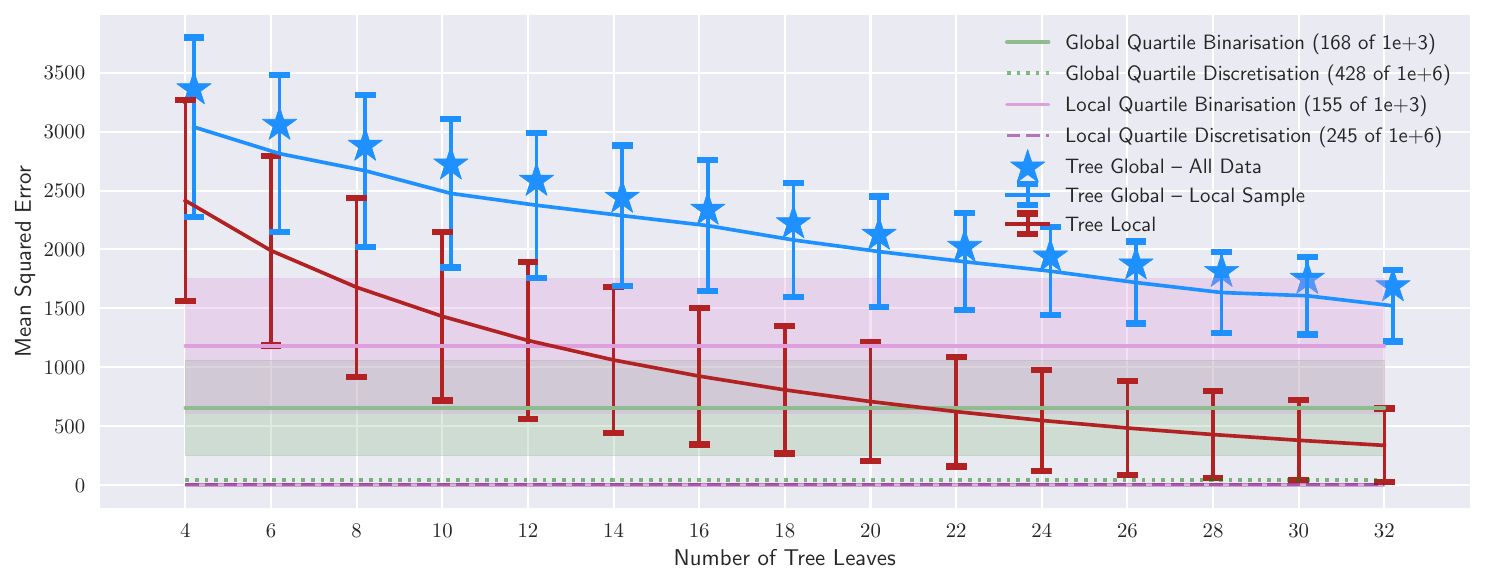}%
        \caption{Weighted average of the \emph{mean squared error} computed for the \emph{diabetes} data IRs.\label{fig:tab_exp:diabetes}}%
    \end{subfigure}
    \caption{%
Interpretable representations based on decision trees achieve better faithfulness of hyper-rectangles (y-axes, lower is better) with fewer encodings (x-axes, small jitter added for readability), i.e., they are more flexible and expressive. %
This property is measured with a weighted average (over IR hyper-rectangles) of Gini impurity for classification tasks and mean squared error for regression or probabilistic classification tasks. %
The number of unique encodings generated by quartile-based IRs is constant for a data set and it is displayed in the legend, shown as the maximum number of encodings used, out of the theoretical limit supported by the IR; %
for tree-based IRs, on the other hand, it is equivalent to the number of leaves, which is recorded on the x-axes. %
Panels~(\subref{fig:tab_exp:boston}) and (\subref{fig:tab_exp:diabetes}) do not capture the tree width at which this IR outperforms the global and local quartile discretisation steps alone, which is 80 or 64 (compared to 441 or 337) and 224 or 112 (compared to 428 or 245) respectively for the \emph{housing} and \emph{diabetes} data sets.%
\label{fig:tab_exp}}%
\end{figure}

\paragraph{Experiments}%
We compare faithfulness of these two tabular interpretable representations on four real-life data sets, two of which are classification and the other two regression problems:%
\begin{itemize}
    \item wine recognition\footnote{\url{https://archive.ics.uci.edu/ml/datasets/wine}} (classification);%
    \item breast cancer Wisconsin diagnostic\footnote{\url{https://archive.ics.uci.edu/ml/datasets/Breast+Cancer+Wisconsin+(Diagnostic)}} (classification);%
    \item Boston house prices\footnote{\url{https://archive.ics.uci.edu/ml/machine-learning-databases/housing}} (regression); and%
    \item diabetes\footnote{\url{https://www4.stat.ncsu.edu/~boos/var.select/diabetes.html}} (regression).%
\end{itemize}
Using these data, we evaluate \emph{quartile-} and \emph{tree-based} IRs in two variants:%
\begin{description}[labelindent=.5em,leftmargin=\parindent+.5em,font=\bfseries]%
  \item [global] where a single discretisation is generated for all the data points (and, in case of the quartile method, followed by creation of instance-specific binary IRs); and%
  \item [local] where a collection of distinct discretisations is composed separately for each individual data point (and then binarised based on the same instance for the quartile method).%
\end{description}
In addition to evaluating the quartile-based binary interpretable representation, we compute faithfulness of the intermediate discretisation step to facilitate an in-detail comparison. %
The results of our experiments are depicted in Figure~\ref{fig:tab_exp}, which reveals that tree-based IRs require a fraction of the expressiveness -- i.e., unique encodings in the binary interpretable space -- used by the quartile-based IRs to achieve a comparable level of hyper-rectangle faithfulness, especially in the \emph{local} variant. %
This can be understood as tree-based interpretable representations being better able to capture the intricacies of the underlying (black-box) labelling mechanism with less complexity to the benefit of the ensuing explanations.%

More precisely, Figure~\ref{fig:tab_exp} displays the impurity of interpretable representations achieved for a range of different tree widths, with the x-axes showing the limit imposed on the number of leaves. %
In each case, the leaves number can be compared to the number of unique hyper-rectangles generated by the discretisation and binarisation steps of the corresponding quartile-based IR. %
The y-axes, on the other hand, depict weighted Gini impurity or mean squared error, respectively for classification and regression tasks, computed for all the hyper-rectangles of each individual IR (Equation~\ref{eq:appendix:tree:score}). %
The dotted green and dashed pink lines labelled as ``global quartile discretisation'' and ``local quartile discretisation'' are the measure of impurity for the quartile discretisation step that underlies this type of an IR. %
The solid green and pink lines surrounded by shading -- marked as ``global quartile binarisation'' and ``local quartile binarisation'' -- correspond to the mean and standard deviation of the hyper-rectangle impurity scores computed for the global and local variants of quartile-based binary IRs (i.e., \emph{discretisation} followed by \emph{binarisation}) for each individual instance in a data set. %
Equivalent measurements are taken for the global and local tree-based IRs for a range of tree widths: ``tree global'' depicted in blue (the \(\star\) symbol corresponds to evaluation on the entire data set whereas the curve captures the same measurement for the neighbourhood of each instance) and ``tree local'' plotted using the red curve, with the error bars denoting the standard deviation. %
In all of the plots, a lower score on the y-axes -- capturing weighted faithfulness of an IR -- is better.%

The pair of numbers placed in brackets next to the quartile discretisation and binarisation labels in the legend of each plot communicates the maximum number of distinct hyper-rectangles for the former, and their binarisation-driven combinations for the latter, that are being used by the validation data, out of all the possible unique values that, respectively, the quartile discretisation and its binarisation can theoretically encode. %
These quantities are directly comparable to the width of trees -- recorded on the x-axes of the plots -- used to partition the feature space to compose the tree-based IRs. %
Given a lack of a black-box model, whose predictions should be used to capture the distribution of the target variable within each hyper-rectangle, we instead utilise the ground truth provided with the aforementioned data sets since this proxy does not affect the validity of our experiments in any way. %
In summary, Figure~\ref{fig:tab_exp} illustrates that interpretable representations created with decision trees are more pure (i.e., uniform) than their quartile-based alternatives, therefore they are superior at capturing the complexity of the underlying labelling mechanism, whatever it may be. %
Furthermore, they achieve better performance with just a fraction of the encodings required by the other method, i.e., they are more expressive because of the elaborate (class-aware) mechanism used by decision trees to partition and merge a feature space.%

\section{Linking Interpretable Representations with Surrogate Models: Analysis of Tabular Data Explainability\label{sec:ols}}%

Interpretable representations are paired with transparent predictive models to form surrogate explainers~\cite{sokol2019blimey}. %
Linear models are a common choice that allows to capture the influence of human-comprehensible concepts on black-box predictions~\cite{friedman2008predictive,ribeiro2016why}, in which case such explanations -- determined by the coefficients of the underlying surrogate linear classifier -- are subject to assumptions and limitations of these models. %
In particular, such explanatory insights can be deceiving when the target variable is \emph{non-linear} with respect to data features, the attributes are \emph{co-dependent} or \emph{correlated}, and the feature values are \emph{not normalised} to the same range~\cite{sokol2019blimey,sokol2020limetree}. %
Intuitively, the first two properties may not hold for high-level interpretable representations since their components are highly inter-dependent -- e.g., adjacent image segments, neighbouring words and bordering hyper-rectangles -- therefore the resulting explanations can misrepresent the possible relations between these concepts and the behaviour of the explained black box. %
\citet{friedman2008predictive} addressed some of these concerns by using logical rules extracted from random forests as the binary interpretable concepts, which they then modelled with a linear predictor; however, the overlap between these rules still violates the feature independence assumption.%

\begin{figure}[t]%
    \centering
    \includegraphics[width=0.3375\textwidth]{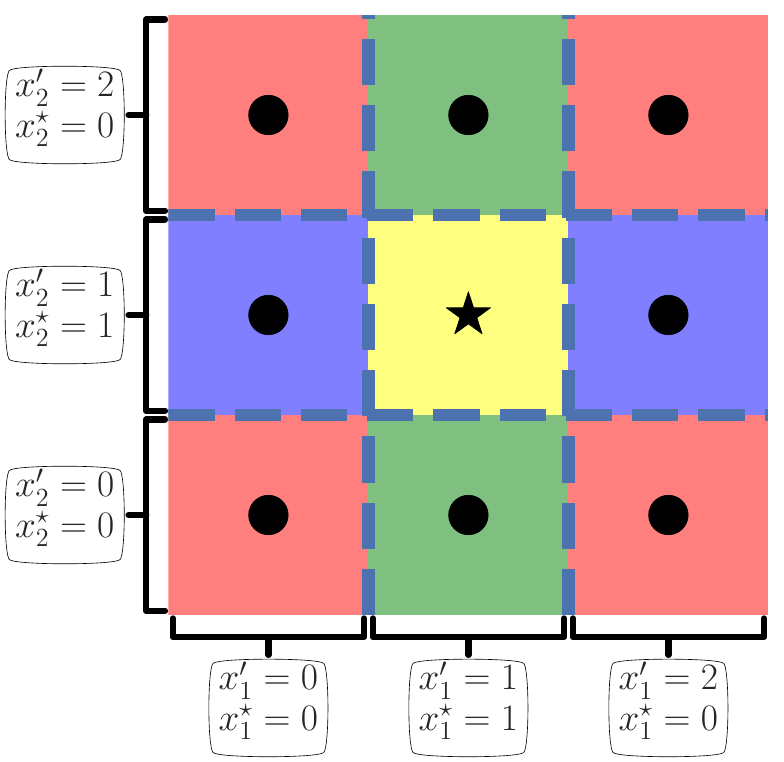}%
    \caption{%
Example of a discrete representation \(x^\prime = (x_1^\prime, x_2^\prime)\) and binary IR \(x^\star = (x_1^\star, x_2^\star)\) of tabular data. %
The \(\star\) symbol represents the explained instance. %
(Refer to Figure~\ref{fig:tabular_binary_ir} for a description of the toy data set used here.) %
\label{fig:tabular_binary_ir:centred}}%
\end{figure}

In addition to these limitations, linear models are inherently \emph{incompatible} with the interpretable representation of tabular data introduced in Section~\ref{sec:ir:tabular}. %
Recall that the information loss suffered when transitioning from the discrete into the binary representation partially forfeits the preceding effort of the discretisation step to faithfully capture the black-box decision boundary. %
The undesired side effect of this procedure adversely affects the weights of the linear model trained on top of such a binary IR. %
This can be observed by deriving an analytical solution to \emph{ordinary least squares} in this specific setting, which is presented in Equation~\ref{eq:importance} for a toy example with two numerical features similar to the scenario shown in Figure~\ref{fig:tabular_binary_ir:centred}. %
In this case, the coefficients \(\Theta_{\mathbf{W}}\) of the OLS model depend on:%
\begin{enumerate}%
    \item %
the number of data points \(w_{ij}\) in the hyper-rectangles determined by the \(x^\star = (i, j)\) coordinates of the binary interpretable representation; and%
    \item %
the average black-box prediction \(\widebar{y}\) in various IR partitions denoted by \(\mathcal{W}_{ij}\), with the set of all the data points given by \(\mathcal{W}\).%
\end{enumerate}%
This formulation can be generalised to an arbitrary number of dimensions spanning numerical and categorical features, and it is applicable to regressors as well as crisp and probabilistic black-box classifiers. %
The derivation of this result is outlined in Appendix~\ref{sec:appendix:ols}.%

\begin{equation}\label{eq:importance}
    \Theta_{\mathbf{W}} =%
{\begin{bNiceMatrix}[%
  columns-width = auto,
  code-after={
    \tikz \node [highlight_blue = (2-2) (3-3)] {} ;
    \tikz\path [fill=red!15, blend mode = multiply, name suffix = -medium, rounded corners = 0.5 mm]
    ($(1-1.north east)+(0pt,1pt)$)
    -- ($(1-3.north east)+(1pt,1pt)$)
    -- ($(1-3.south east)+(1pt,-1pt)$)
    -- ($(1-1.south east)+(1pt,-1pt)$)
    -- ($(3-1.south east)+(1pt,-1pt)$)
    -- ($(3-1.south west)+(-1pt,-1pt)$)
    -- ($(1-1.north west)+(-1pt,1pt)$)
    -- cycle ;
  }]
1 & \frac{w_{11} + w_{10}}{\sum{w_{ij}}} & \frac{w_{11} + w_{01}}{\sum{w_{ij}}}\\
1 & 1 & \frac{w_{11}}{w_{11} + w_{10}}\\
1 & \frac{w_{11}}{w_{11} + w_{01}} & 1
\end{bNiceMatrix}}^{-1}
\times
\begin{bNiceMatrix}[%
  code-after={
    \tikz \node [highlight_red = (1-1)] {} ;
    \tikz \node [highlight_blue = (2-1) (3-1)] {} ;
  }]
\widebar{y}_{\mathcal{W}}\\
\widebar{y}_{\mathcal{W}_{11}\cup\mathcal{W}_{10}}\\
\widebar{y}_{\mathcal{W}_{11}\cup\mathcal{W}_{01}}
\end{bNiceMatrix}
\end{equation}

This outcome allows us to draw conclusions about the meaning of the interpretable concept influence given by the coefficients of a linear surrogate when the intercept is modelled (red and blue shading in Equation~\ref{eq:importance}), and without it (blue shading). %
In particular, the influence of interpretable concepts is \emph{solely} based on:%
\begin{itemize}
    \item \textbf{the proportion determined by the number of the data points} residing in the explained hyper-rectangle (\(\mathcal{W}_{11}\)) divided by the count of points located in the hyper-rectangles aligned with the explained hyper-rectangle along every axis, i.e., \(\mathcal{W}_{11} \cup \mathcal{W}_{10}\) for the first feature and \(\mathcal{W}_{11} \cup \mathcal{W}_{01}\) for the second attribute; and%
    \item \textbf{the average value predicted by the explained black box} in the latter two subspaces -- \(\mathcal{W}_{11} \cup \mathcal{W}_{10}\) and \(\mathcal{W}_{11} \cup \mathcal{W}_{01}\) -- scaled appropriately when the intercept is modelled.%
\end{itemize}
For example, consider Figure~\ref{fig:tabular_binary_ir:centred} where \(x_1^\star\) denotes the first binary interpretable feature and \(x_2^\star\) the second. %
In this case, \(\mathcal{W}_{11}\) is the yellow hyper-rectangle; \(\mathcal{W}_{11} \cup \mathcal{W}_{10}\) is the union of the yellow and green hyper-rectangles; and \(\mathcal{W}_{11} \cup \mathcal{W}_{01}\) is the combination of yellow and blue hyper-rectangles. %
Finally, \(\widebar{y}_{\mathcal{W}_{11}\cup\mathcal{W}_{10}}\) is the average prediction in the vertical green and yellow column, and \(\widebar{y}_{\mathcal{W}_{11}\cup\mathcal{W}_{01}}\) is the average prediction in the horizontal blue and yellow row.%

When modelled, the intercept value is additionally determined by:%
\begin{itemize}
    \item \emph{the proportion} given by the number of data points in the hyper-rectangles aligned with the explained hyper-rectangle along every axis, divided by the total number of data points; and%
    \item \emph{the average} value predicted by the black box for all the data points.%
\end{itemize}
Intuitively, the instances not aligned with the explained hyper-rectangle -- the red blocks in Figure~\ref{fig:tabular_binary_ir:centred} -- are assigned the \(x^\star = (0, 0)\) coordinates in the binary interpretable representation, therefore they cannot contribute to the feature coefficients of a linear model, just the intercept. %
This can be easily seen with the \(g(x^\star; \Theta) = \sum_{i=0}^n \Theta_i x^\star_i\) formula, where \(x^\star_0 = 1\) is the phantom feature and the remaining data features \(x^\star_1,\dots,x^\star_n\) are all \(0\); therefore, these instances can only influence the intercept coefficient \(\Theta_0\).%

An important insight uncovered by our results is \textbf{partial insignificance of the discretisation quality} given a fixed number of data points placed in the identified collections of relevant hyper-rectangles. %
Using this property we can \emph{manipulate} the explanation by altering the number of data points in meaningful partitions, with the discretisation faithfulness having relatively minor influence. %
For example, consider the two discretisations depicted earlier in Figure~\ref{fig:tabular_ir}, assuming that the explained hyper-rectangle is \(x^\prime = (1, 1)\) for both panels, and that the \(x^\prime = (1, 0)\) and \(x^\prime = (1, 1)\) partitions in Figure~\ref{fig:tabular_ir:class} have \emph{three} additional data points each. %
In this scenario, when modelling the influence of interpretable components without the intercept, the only difference between these two cases are the black-box predictions of the instances placed in the expanded hyper-rectangles, i.e., \(x^\prime = (1, 0)\) and \(x^\prime = (1, 1)\), since:%
\begin{description}[labelindent=.5em,leftmargin=\parindent+.5em,font=\bfseries]%
    \item [Figure~\ref{fig:tabular_ir:dist}] \(w_{1 1} = 4\), \(w_{0 1} = 4 + 4 = 8\) and \(w_{1 0} = 4\), leading to \(\frac{w_{1 1}}{w_{1 1} + w_{1 0}} = \frac{4}{4 + 4} = \frac{1}{2}\) and \(\frac{w_{1 1}}{w_{1 1} + w_{0 1}} = \frac{4}{4 + 8} = \frac{1}{3}\); and%
    \item [Figure~\ref{fig:tabular_ir:class}] \(w_{1 1} = 2 + \mathit{3} = 5\), \(w_{0 1} = 4 + 4 + 2 = 10\) and \(w_{1 0} = 2 + \mathit{3} = 5\), leading to \(\frac{w_{1 1}}{w_{1 1} + w_{1 0}} = \frac{5}{5 + 5} = \frac{1}{2}\) and \(\frac{w_{1 1}}{w_{1 1} + w_{0 1}} = \frac{5}{5 + 10} = \frac{1}{3}\).%
\end{description}

Depending on the gradient smoothness of the underlying probabilistic black box, these explanations may slightly differ. %
However, if the additional six data points are placed such that the average black-box predictions of \(\mathcal{W}_{11} \cup \mathcal{W}_{10}\) and \(\mathcal{W}_{11} \cup \mathcal{W}_{01}\) are identical across both discretisations, the resulting explanations will be the same. %
Alternatively, if they are crisp predictions instead of class probabilities, the two explanations will also be indistinguishable regardless of where the additional six instances are situated within their respective hyper-rectangles. %
Note that in general it is easier to manipulate the explanations when dealing with crisp predictions rather than probabilities as we only have to consider which side of the black-box decision surface -- if one runs across a given hyper-rectangle -- to place each data point. %
The added benefit of this observation is evidence that discretising each numerical feature into more than three bins is not necessarily beneficial, with the most important partition boundaries being the ones enclosing the explained data point.%
\footnote{To facilitate further exploration of the explanatory setting discussed in this section we implemented a no-code interactive widget within a Jupyter Notebook and published it on GitHub at \url{https://github.com/fat-forensics/resources/tree/master/tabular_surrogate_builder}. It allows to investigate the influence of the numerical feature discretisation and the number of data points placed in each hyper-rectangle on the surrogate explanation extracted from a linear model by manually adjusting these parameters.}%

This complex relation between explanations, the discretisation underlying an interpretable representation and the distribution of data points transformed by the IR to fit a surrogate linear model must be well-understood to ensure the veracity of explanatory insights. %
Given how sensitive an explanation is to these factors, small variations to the parameterisation of the aforementioned building blocks may sometimes yield disparate or even opposing insights into the behaviour of a black box. %
Such counterintuitive explanations can, for example, be achieved by shifting the discretisation boundaries for a fixed data sample, or instead by moving around these instances with a fixed IR. %
The discretisation process should therefore be optimised to guarantee the most truthful explanation that cannot be easily swayed by altering the data sample either in its distribution or size.%

To overcome some of these problems and facilitate explanations that are more diverse than influence of interpretable concepts, alternative surrogate models can be used~\cite{sokol2019blimey}. %
Logical predictors, such as \emph{decision trees}, are particularly appealing given that they provide a wide range of explanations and they do not introduce any restrictions on the behaviour or interrelation of features, albeit they do impose axis-parallel partition of the feature space~\cite{sokol2020limetree,sokol2021towards}. %
They are particularly suited for explaining tabular data, for which they alleviate the need for a separate interpretable representation as noted in the previous section. %
In particular, they can automatically learn a locally faithful, class-aware discretisation, with the added benefit of modelling combinations of hyper-rectangles and not suffering from information loss or stochasticity when applying the IR transformation~\cite{sokol2019blimey}. %
In the following section we explore more guidelines that can help us to design and build robust and trustworthy interpretable representations with well-understood properties.%

\section{Towards Robust and Trustworthy Interpretable Representations\label{sec:guidelines}}

Our investigation of interpretable representations has revealed that a one-size-fits-all approach is often suboptimal. %
As with many other steps of the machine learning workflow, IRs need to be crafted for the problem at hand to be useful, reliable, robust and trustworthy~\cite{rudin2019stop,sokol2021explainability}. %
Moreover, the way in which an interpretable representation is built and operationalised determines the meaning of the resulting explanations in addition to constraining their possible types and compatible communication media. %
IR properties and desiderata should therefore be well-understood, guiding their development and deployment in each unique context. %
By following best practice -- aspects of which are outlined below -- we can improve veracity and faithfulness of post-hoc and model-agnostic explainers that rely on IRs, thus address some criticism of these techniques~\cite{rudin2019stop}. %
Notably, consulting the latest findings in each relevant discipline -- natural language processing, computer vision and discretisation of numerical data -- can offer a treasure trove of insights and contribute core concepts to the fundamental design of interpretable representations. %
Popular data pre-processing and feature engineering techniques used across these domains can be adapted to build more informative IRs; for example, word clustering for text data, object detection and semantic segmentation for images, and subgroup discovery for tabular data. %

\paragraph{Information Loss}%
While the many-to-one mapping pertinent to tabular IRs may seem detrimental at first due to the resulting loss of information, this procedure creates sparsity and reduces the perceived complexity of the data, which are often a prerequisite of human intelligibility~\citep{sokol2023unreasonable}. %
This situation is unique to tabular data since both images and text are inherently comprehensible. %
Given the necessity of representing numerical features as human-understandable categories, the mapping should focus on eliciting the most insightful concepts and only discard redundant information (refer to Section~\ref{sec:disc:tab}); for example, partitioning a range of numbers into relatively uniform bins with respect to the underlying labels while ensuring that they are also meaningful to explainees. %
Notably, non-adjacent numerical intervals can be combined into a single interpretable concept if such an aggregation improves human understanding. %
The discretisation process should be driven by precise optimisation and evaluation objectives that are defined based on the explanatory context, domain constraints and user expectations. %
Since different thresholds can yield distinct or even opposing explanations -- thus adversely impacting their trustworthiness -- it is important to set out well-defined goals and metrics that capture these properties in detail.%

\paragraph{Optimisation and Evaluation Criteria}%
Interpretable representation desiderata should be formalised to allow for straightforward IR optimisation, testing and comparison both with respect to domain-specific characteristics and technical properties, thus precisely guiding the development and evaluation of IRs. %
While the former objective is difficult to define in an application-agnostic way, the latter should take into consideration the structure of the entire data or the specific neighbourhood being explained (refer to Section~\ref{sec:disc:tab}). %
Since human comprehension of text is intrinsically consistent with how the corresponding IRs are operationalised -- i.e., switching concepts on and off by removing relevant words from an excerpt of text -- these criteria appear to be entirely encapsulated by the user's perception of individual tokens, which largely depends on the underlying pre-processing step (refer to Section~\ref{sec:ir:text}). %
A similar line of reasoning applies to images, regardless of whether the IR is based on edge detection or semantic segmentation; the objective is to separate visual concepts that are distinct from a human perspective and relevant to the predictive task being explained (refer to Sections~\ref{sec:ir:images} and \ref{sec:appendix:img}). %
For example, an algorithmically generated image IR can be improved by merging, possibly non-adjacent, super-pixels representing the background of an object into a single interpretable concept.%

Tabular data, on the other hand, in themselves provide a rich source of information that can be used to algorithmically navigate the discretisation process that underpins the corresponding interpretable representation. %
Characteristics such as data density, distribution, their black-box predictions and confidence thereof can be used to partition a feature space into geometrically consistent and homogeneous concepts. %
Generic metrics that determine the purity, uniformity and faithfulness of discretised data can be utilised to this end; for example, see the evaluation strategy proposed in Section~\ref{sec:disc:tab}. %
While our analysis showed that discretising each numerical attribute into more than three bins is not necessarily beneficial (refer to Section~\ref{sec:ols}), this is purely a consequence of the hyper-rectangle merging procedure employed by the binarisation step that follows. %
By designing a more complex transformation function from the discrete space into the binary interpretable representation, the IR can benefit from higher discretisation granularity, e.g., consider allowing multiple hyper-rectangles to contribute to the explained concept. %
Additional improvements can include only using intervals that are bounded from each side to narrow down the scope of the explanation and prevent it from being biased by out-of-distribution instances. %
The optimisation and evaluation of tabular IR discretisation can further be enhanced by domain-specific knowledge that assigns a human-comprehensible concept to each partition, akin to how image segmentation may be assessed based on the semantic consistency of visual object separation. %
Finally, in addition to an independent validation of IR quality, the robustness and stability of the resulting explanations can be measured while varying IR parameters~\citep{sokol2020tut}.%

\paragraph{Human-in-the-loop Design}%
Explainee-driven interactive creation or personalisation of interpretable representations is an interesting avenue of research on the crossroads of explainable artificial intelligence and human--computer interaction~\cite{sokol2018glass,sokol2020one,lage2020human}. %
It has the potential to formulate further recommendations for the composition and operationalisation of IRs for individual applications, but such a solution comes at the expense of a user-in-the-loop architecture that may be difficult to automate and scale. %
This design choice, however, can be easily justified since constructing an interpretable representation that is intelligible and useful is often user- and application-dependent or even unique to the explained data point. %
Moreover, the core premise of IRs is to encode concepts that are \emph{meaningful} to the target audience and \emph{relevant} to the question that prompted the need for explainability in the first place, thus relying upon computer-generated IRs without communicating their behaviour and properties to the explainees may be counterproductive (refer to Section~\ref{sec:appendix:img}). %
While promising, scaling up human-in-the-loop interpretable representations appears to be impractical without a concrete deployment use case.%

\paragraph{Information Removal (Proxy)}%
The interpretable representations that we deal with in this paper capture human-intelligible concepts that can be switched on or off. %
This process defines the explanatory \emph{fact} and \emph{foil}, the distinction between which forms the basis of insights into the predictive behaviour of the model under investigation. %
Given the significance of this procedure, the difference between the two should be \emph{semantically meaningful} as well as \emph{computationally effective} in the chosen operational context (refer to Section~\ref{sec:ir-analysis}). %
To this end, the strategy used to discard information from the explained instance must achieve its goal and avoid introducing unintended biases, both of which can be measured by observing the response of a black box when predicting data manipulated via an interpretable representation. %
This process does not affect text since its IRs support a direct removal of tokens and relevant predictive models can handle instances that have been altered in this way, however images and tabular data require an algorithmic proxy. %

For images, this is achieved through \emph{content replacement}, with retouching, object injection and region occlusion being the most popular choices (see Section~\ref{sec:ir:images} for more details). %
While the evidence presented in Section~\ref{sec:appendix:img} clearly shows the disadvantage of using mean-colour occlusion and a relatively comparable properties of all the other tested colouring strategies, our results may not generalise to different image data sets and black-box models, therefore a similar analysis should be performed prior to deciding on the content replacement technique. %
Even if such an investigation shows that, overall, a selected information removal proxy behaves comparably to others for a specific setup, it may still be inappropriate for a particular image and explained class. %
For example, the \emph{white} and \emph{black} occlusion strategies are nearly indistinguishable in our experiments (refer to Figure~\ref{fig:img_exp}), but employing the former for the dog image used across this paper for illustration purposes (shown in Figure~\ref{fig:img}) may be ineffective for explaining it, and \emph{winter scenery} in general, given the photo's snowy background. %
Therefore, unless the data domain is highly homogeneous -- e.g., all pictures follow a fixed object presentation pattern and colour scheme -- the information removal proxy may need to be configured on a case-by-case basis. %

An information removal proxy for tabular data is more complex given that it operates on an interpretable representation that is based on discretisation of numerical attributes followed by a binarisation step. %
Specifically, switching off an IR element is equivalent to placing the value of the corresponding numerical feature outside of the range encoded by this concept; or, if the attribute is categorical, choosing any other value not captured by this concept. %
Therefore, this operation should ensure that moving data between different hyper-rectangles (or their collections) determined by distinct binary interpretable spaces is \emph{semantically meaningful} and corresponds to abstractions that can be captured \emph{computationally} (e.g., by monitoring the change of black-box predictions for these instances).%

When working with categories generated via discretisation, a step in this direction can be a more informed process of merging these hyper-rectangles into binary concepts -- in contrast to doing so based on their geometrical alignment -- in addition to narrowing down the scope of the foil by explicitly bounding the numerical ranges instead of comparing spaces that cannot be easily represented by finite data samples. %
A similar strategy may also be applied to images and text, where (non-adjacent) super-pixels and word-based tokens can be combined into a single IR component. %
Additionally, with a well-crafted tabular interpretable representation, the volatility of explanations may be reduced since such an IR becomes less reliant on the distribution of the (possibly random) data sample transformed into this domain and used to train a local surrogate model (refer to Section~\ref{sec:ols}). %
Nonetheless, building an IR that expresses the fact and the foil as complementary events and allows to intuitively manipulate them by tweaking each IR component \emph{independently} may be impractical or impossible to achieve, in which case self-contained regions that do not directly rely on manipulation of individual feature values -- e.g., determined by decision tree leaves or data clustering -- offer an attractive alternative.%

\paragraph{Stochasticity}%
Surrogate explainers -- a big beneficiary of interpretable representations -- sample the data required to train the (local) model directly from the binary IR~\cite{ribeiro2016why}. %
While reasonable for images and text where generating data by manipulating raw pixels, letters or words may easily introduce inconsistencies, following the same procedure for a tabular IR entails converting the binary sample back into the original domain (to be predicted by the explained black box), which requires \emph{random sampling} because of the many-to-one forward transformations -- see Figure~\ref{fig:tab} for reference.%
\footnote{%
Transitioning from an original into a discrete representation is a many-to-one operation if the underlying data set contains numerical features. %
Transforming the discrete representation into a binary IR is also a many-to-one mapping if any discretised attribute has more than two unique values. %
Section~\ref{sec:disc:tab} discusses the information loss pertinent to tabular interpretable representations in more detail.%
} %
Specifically, to execute this process we first choose \emph{at random} one of the merged hyper-rectangles if the binary component is \(0\); \(1\) uniquely identifies a hyper-rectangle in our case. %
Next, we draw a numerical value from the range defined by this hyper-rectangle, e.g., using a (truncated) Gaussian distribution fitted to the (training) data enclosed by this hyper-rectangle; categorical features are uniquely identified by a hyper-rectangle. %
However, tabular data can be sampled in their original representation, which removes the need for this stochastic operation, thus improving robustness and decreasing volatility of surrogate explanations~\cite{sokol2019blimey}.%

Data drawn from the original domain can be easily transformed into a discrete representation and then binarised. %
Moving in the opposite direction in a deterministic fashion requires memorising the correspondence between these points in different representations when executing the forward transformation. %
This matching offers an algorithmic workaround that can be compared to storing the pixel structure and segment adjacency for images or a sentence skeleton and any pre-processing steps for text. %
By sampling in the original domain and connecting different representations of each instance we avoid using the stochastic inverse IR transformation of tabular data, hence reduce randomness and improve stability of the resulting explanations. %
However, this strategy forfeits the implicit locality and diversity achieved by operating directly on the binary representation, therefore the substitute sampling algorithm should directly target a well-defined subspace to carefully capture the behaviour of the explained black box in this region~\cite{sokol2019blimey}. %
More broadly, recognising the strengths and weaknesses of individual components from which explainability algorithms are built allows us to adapt their architecture accordingly, creating the best possible explainer for the problem at hand~\cite{sokol2020tut}.%

\paragraph{Alternative Surrogate Models}%
Interpretable representations offer a sparse and human-comprehensible medium for communicating explanations of black-box models and their predictions. %
Section~\ref{sec:ols}, nonetheless, demonstrated that certain pairings of IRs and surrogate models yield defective explainers whose insights can be misleading or outright incorrect. %
Given the data pre-processing that interpretable representations entail, \emph{logical predictive models} appear to be a good (surrogate model) candidate since they are inherently transparent and intelligible. %
In particular, rule lists and decision trees should be considered to this end, with the latter choice being especially appealing for tabular data for which they can automatically compose an IR in addition to modelling it~\citep{sokol2019blimey}. %
Moreover, they account for feature interactions, and their optimisation procedure is well-aligned with the IR faithfulness objectives discussed in Section~\ref{sec:disc:tab}. %
While tree training procedures tend to be greedy, alternatives that consider multiple features at any given iteration could improve the quality of the resulting interpretable representations and surrogate explainers even further.%

\section{Conclusion and Future Work\label{sec:conclusion}}%

Our findings show the importance of building robust, trustworthy and algorithmically sound interpretable representations as well as their role in defining the question answered by the resulting explanations and veracity thereof. %
Among others, we demonstrated that building IRs with generic algorithms may lead to subpar explainers, and that the intended application domain and audience should always be accounted for, in addition to considering interactive customisation and personalisation of interpretable representations. %
In particular, we discussed a popular operationalisation of IRs for image, text and tabular data in which they are used as binary indicators of presence and absence of human-intelligible concepts. %
This framework is then combined with surrogate models to quantify the influence of such concepts on individual black-box predictions.%

In this setting, we identified challenges such as implicit assumptions, flawed information removal proxies, undesired parametrisation choices, insufficient faithfulness and transformation stochasticity -- which are particularly prominent for tabular and image data -- and showed how to overcome them. %
We also demonstrated the limitations of explaining binary interpretable representations of tabular data with linear models and suggested logical models as a viable alternative. %
Our findings reinforce the importance of considering the structure of the data, especially in the neighbourhood of the explained instance, when transforming them into an IR as well as having a well-defined objective and evaluation metric to aid in IR construction and optimisation. %
Many of these goals can be achieved by drawing inspiration from the fields of computer vision, natural language processing and data discretisation, and more broadly data wrangling and modelling in machine learning, which can inform better design of interpretable representations.%

Our future work will investigate algorithmic information removal proxies, focusing on meaningful and effective occlusion approaches for images and feature value replacement techniques for tabular data; %
we will also look into user-in-the-loop design of interpretable representations. %
Furthermore, we will survey inherently transparent models that are best suited to various IRs, ensuring their technological compatibility and explanatory appeal, in particular focusing on different types of logical predictive models. %
To enable safe adoption of interpretable representations in real-life applications we will finally evaluate the most promising approaches with targeted user studies.%

\renewcommand{\acksname}{Acknowledgements}
\begin{acks}
This work was supported by the TAILOR Network -- an ICT-48 European AI Research Excellence Centre funded by EU Horizon 2020 research and innovation programme (grant agreement number 952215). %
Additional funding was provided by the %
ARC Centre of Excellence for Automated Decision-Making and Society, sponsored by the Australian Government through the Australian Research Council (project number CE200100005); and %
the Hasler Foundation (grant number 23082). %
\end{acks}

\section*{Conflicts of Interest}%
We declare that we have no conflicts of interest.%

\section*{Data Transparency}
Not applicable.%

\section*{Code Availability}
All of the experimental results presented in this paper can be reproduced with FAT Forensics~\cite{sokol2020fat,sokol2019fat} -- a Python package that offers modular implementations of various explainability algorithms -- with the experiment code available on GitHub\footnote{\url{https://github.com/So-Cool/bLIMEy/tree/master/DAMI_2024}}. %
Additional resources -- such as computational notebooks -- that explore properties of tabular and image interpretable representations in the context of surrogate explainers are available as part of the FAT Forensics documentation\footnote{\url{https://fat-forensics.org/how_to/index.html}} and the ``What and How of Machine Learning Transparency'' hands-on tutorial\footnote{\url{https://github.com/fat-forensics/Surrogates-Tutorial}} organised at ECML-PKDD 2020~\cite{sokol2020tut}.%

\bibliographystyle{ACM-Reference-Format}
\bibliography{ir}

\clearpage\newpage

\appendix

\section{Derivation of OLS Explanations for Binary IRs of Tabular Data\label{sec:appendix:ols}}%

When analysing the behaviour of algorithmic black boxes with surrogate explainers, linear models can be used to quantify the positive or negative influence of interpretable concepts extracted from a data point of interest on its black-box prediction~\cite{friedman2008predictive,ribeiro2016why,sokol2019blimey}. %
For some binary interpretable domains, however, such an approach is inherently flawed; %
it is particularly problematic for tabular data transformed into the binary interpretable representation introduced in Section~\ref{sec:ir:tabular}, i.e., achieved through feature discretisation followed by a binarisation step. %
In this appendix, we derive a closed-form expression of this explanation type, which allows us to %
demonstrate how the influence of an interpretable concept measured by the coefficients of a linear model may be deceiving. %
The insights stemming from our analysis can be used to manipulate surrogate explanations, e.g., those produced by LIME~\cite{ribeiro2016why}, through a specially crafted, yet perfectly valid, IR discretisation and data sample.%

Our results are based on the analytical solution to unweighted (\(\Theta\)) and weighted (\(\Theta_{\mathbf{W}}\)) Ordinary Least Squares outlined in Equations~\ref{eq:appendix:ols} and \ref{eq:appendix:ols_weighted} respectively, where \(\mathbf{W}\) is the weight matrix, \(\mathbf{X}\) is the binary interpretable representation data matrix, and \(\mathbf{y}\) is a vector holding the corresponding black-box predictions. %
In our analysis, we assume that the model under investigation is a probabilistic classifier, in which case \(\mathbf{y}\) captures probabilities of the explained class; nonetheless, a similar line of reasoning applies to regressors and crisp classifiers. %
In the latter scenario, the elements of \(\mathbf{y}\) are assumed to be \(1\) when the black-box predictions are the same as the explained class, and \(0\) for any other class. %
Modelling \(\mathbf{y}\) in such a way generates one-vs-rest explanations -- i.e., evidence for the black box predicting the explained rather than any other class -- akin to the insights produced for probabilistic black boxes, for which the surrogate is trained \emph{only} on the probabilities of the \emph{explained class}. %
Therefore, both approaches measure the influence of interpretable concepts -- determined by the coefficients of the linear model -- on a selected class when tasked with telling it apart from all the other classes.%

\begin{gather}%
    \label{eq:appendix:ols}
    \Theta = (\mathbf{X}^T\mathbf{X})^{-1} \; \mathbf{X}^T\mathbf{y} \\
    \label{eq:appendix:ols_weighted}
    \Theta_{\mathbf{W}} = (\mathbf{X}^T\mathbf{W}\mathbf{X})^{-1} \; \mathbf{X}^T\mathbf{W}\mathbf{y}
\end{gather}%

In the interest of brevity and readability, we analyse tabular data with two numerical features -- similar to the examples shown in Figures~\ref{fig:tabular_binary_ir} and \ref{fig:tabular_binary_ir:centred} -- nonetheless our findings generalise to an arbitrary number of attributes that are both categorical and numerical. %
In a generic setting, for \(n\) features there will be \(n\) binary concepts with \(2^n\) unique encodings in the interpretable representation (the cardinality thereof). %
If additionally we choose to model the intercept of the linear regression, a phantom all-\(1\) column vector is inserted at the front of the data matrix \(\mathbf{X}\). %
Therefore, the \(\mathbf{X}^T\mathbf{X}\) and \(\mathbf{X}^T\mathbf{W}\mathbf{X}\) components of \(\Theta\) and \(\Theta_{\mathbf{W}}\) respectively are square matrices of \(n \times n\) shape sans the intercept or \((n+1) \times (n+1)\) when the intercept is modelled.%

Figure~\ref{fig:tabular_binary_ir} depicts a simplistic view of data sampling for two numerical attributes with just one instance in each discrete hyper-rectangle. %
In reality, however, we should expect their large quantity since it allows to better approximate the behaviour of the underlying black box, especially when the number of features is high. %
In this particular case, the binary interpretable representation data matrix \(\mathbf{X}\) -- with the first column (red) inserted to model the intercept and the remaining columns (blue) representing the binary data -- is:%
\[%
\mathbf{X} = %
\begin{bNiceMatrix}[%
  code-after={
    \tikz \node [highlight_red = (1-1) (9-1)] {} ;
    \tikz \node [highlight_blue = (1-2) (9-3)] {} ;
  }]
1 & 1 & 1\\
1 & 1 & 0\\
1 & 1 & 0\\
1 & 0 & 1\\
1 & 0 & 1\\
1 & 0 & 0\\
1 & 0 & 0\\
1 & 0 & 0\\
1 & 0 & 0
\end{bNiceMatrix}
\textnormal{,}
\]
which gives:%
\begin{gather*}
\mathbf{X}^T\mathbf{X} = %
\begin{bNiceMatrix}[%
  code-after={
    \tikz \node [highlight_red = (1-1) (1-9)] {} ;
    \tikz \node [highlight_blue = (2-1) (3-9)] {} ;
  }]
1 & 1 & 1 & 1 & 1 & 1 & 1 & 1 & 1\\
1 & 1 & 1 & 0 & 0 & 0 & 0 & 0 & 0\\
1 & 0 & 0 & 1 & 1 & 0 & 0 & 0 & 0
\end{bNiceMatrix}
\times
\begin{bNiceMatrix}[%
  code-after={
    \tikz \node [highlight_red = (1-1) (9-1)] {} ;
    \tikz \node [highlight_blue = (1-2) (9-3)] {} ;
  }]
1 & 1 & 1\\
1 & 1 & 0\\
1 & 1 & 0\\
1 & 0 & 1\\
1 & 0 & 1\\
1 & 0 & 0\\
1 & 0 & 0\\
1 & 0 & 0\\
1 & 0 & 0
\end{bNiceMatrix}
=
\begin{bNiceMatrix}[%
  columns-width = auto,
  code-after={
    \tikz \node [highlight_blue = (2-2) (3-3)] {} ;
    \tikz\path [fill=red!15, blend mode = multiply, name suffix = -medium, rounded corners = 0.5 mm]
    ($(1-1.north east)+(0pt,1pt)$)
    -- ($(1-3.north east)+(1pt,1pt)$)
    -- ($(1-3.south east)+(1pt,-1pt)$)
    -- ($(1-1.south east)+(1pt,-1pt)$)
    -- ($(3-1.south east)+(1pt,-1pt)$)
    -- ($(3-1.south west)+(-1pt,-1pt)$)
    -- ($(1-1.north west)+(-1pt,1pt)$)
    -- cycle ;
  }]
9 & 3 & 3\\
3 & 3 & 1\\
3 & 1 & 3
\end{bNiceMatrix}
\textnormal{.}
\end{gather*}

Since some of the hyper-rectangles are merged when transitioning from the discrete into the binary interpretable representation, \(\mathbf{X}\) contains duplicated rows. %
The influence of this phenomenon is magnified even further when multiple data points are placed within a single hyper-rectangle. %
Without loss of generality, we can use the \emph{weighted} variant of OLS with the data set \(\mathbf{X}\) composed of only one copy of each unique binary data point and the weights corresponding to their counts. %
In this case:%
\[
\mathbf{X} = %
\begin{bNiceMatrix}[%
  code-after={
    \tikz \node [highlight_red = (1-1) (4-1)] {} ;
    \tikz \node [highlight_blue = (1-2) (4-3)] {} ;
  }]
1 & 1 & 1\\
1 & 1 & 0\\
1 & 0 & 1\\
1 & 0 & 0
\end{bNiceMatrix}
\;\;\;\textnormal{and}\;\;\;
\mathbf{W} = %
\begin{bmatrix}
w_{11} & 0 & 0 & 0\\
0 & w_{10} & 0 & 0\\
0 & 0 & w_{01} & 0\\
0 & 0 & 0 & w_{00}
\end{bmatrix}
\textnormal{,}
\]
where \(w_{i j}\) is the count of data points residing in all of the hyper-rectangles that are assigned the \((i, j)\) coordinates in the binary interpretable representation -- see the \((x_1^\star, x_2^\star)\) coordinates in Figure~\ref{fig:tabular_binary_ir} for reference. %
Therefore, for an arbitrary number of instances in a data set with two numerical features when modelling the intercept:%
\begin{align*}
\mathbf{X}^T\mathbf{W}\mathbf{X} =& %
\begin{bNiceMatrix}[%
  code-after={
    \tikz \node [highlight_red = (1-1) (1-4)] {} ;
    \tikz \node [highlight_blue = (2-1) (3-4)] {} ;
  }]
1 & 1 & 1 & 1\\
1 & 1 & 0 & 0\\
1 & 0 & 1 & 0
\end{bNiceMatrix}
\times
\begin{bmatrix}
w_{11} & 0 & 0 & 0\\
0 & w_{10} & 0 & 0\\
0 & 0 & w_{01} & 0\\
0 & 0 & 0 & w_{00}
\end{bmatrix}
\times
\begin{bNiceMatrix}[%
  code-after={
    \tikz \node [highlight_red = (1-1) (4-1)] {} ;
    \tikz \node [highlight_blue = (1-2) (4-3)] {} ;
  }]
1 & 1 & 1\\
1 & 1 & 0\\
1 & 0 & 1\\
1 & 0 & 0
\end{bNiceMatrix}\\
=&
\begin{bNiceMatrix}[%
  code-after={
    \tikz \node [highlight_red = (1-1) (1-4)] {} ;
    \tikz \node [highlight_blue = (2-1) (3-4)] {} ;
  }]
w_{11} & w_{10} & w_{01} & w_{00}\\
w_{11} & w_{10} & 0 & 0\\
w_{11} & 0 & w_{01} & 0
\end{bNiceMatrix}
\times
\begin{bNiceMatrix}[%
  code-after={
    \tikz \node [highlight_red = (1-1) (4-1)] {} ;
    \tikz \node [highlight_blue = (1-2) (4-3)] {} ;
  }]
1 & 1 & 1\\
1 & 1 & 0\\
1 & 0 & 1\\
1 & 0 & 0
\end{bNiceMatrix}\\
=&
\begin{bNiceMatrix}[%
  columns-width = auto,
  code-after={
    \tikz \node [highlight_blue = (2-2) (3-3)] {} ;
    \tikz\path [fill=red!15, blend mode = multiply, name suffix = -medium, rounded corners = 0.5 mm]
    ($(1-1.north east)+(0pt,1pt)$)
    -- ($(1-3.north east)+(1pt,1pt)$)
    -- ($(1-3.south east)+(1pt,-1pt)$)
    -- ($(1-1.south east)+(1pt,-1pt)$)
    -- ($(3-1.south east)+(1pt,-1pt)$)
    -- ($(3-1.south west)+(-1pt,-1pt)$)
    -- ($(1-1.north west)+(-1pt,1pt)$)
    -- cycle ;
  }]
\sum{w_{ij}}         & w_{11} + w_{10} & w_{11} + w_{01}\\
w_{11} + w_{10} & w_{11} + w_{10} & w_{11}         \\
w_{11} + w_{01} & w_{11}          & w_{11} + w_{01}
\end{bNiceMatrix}
\textnormal{.}
\end{align*}
For the example in Figure~\ref{fig:tabular_binary_ir} -- where \(w_{11} = 1\), \(w_{10} = 2\), \(w_{01} = 2\) and \(w_{00} = 4\) -- a calculation for the weighted variant agrees with the previous result computed directly for \(\mathbf{X}^T\mathbf{X}\).%

Next, we analyse the second component of the \(\Theta_{\mathbf{W}}\) formula:%
\begin{align*}
\mathbf{X}^T\mathbf{W}\mathbf{y} =& %
\begin{bNiceMatrix}[%
  code-after={
    \tikz \node [highlight_red = (1-1) (1-4)] {} ;
    \tikz \node [highlight_blue = (2-1) (3-4)] {} ;
  }]
1 & 1 & 1 & 1\\
1 & 1 & 0 & 0\\
1 & 0 & 1 & 0
\end{bNiceMatrix}
\times
\begin{bmatrix}
w_{11} & 0 & 0 & 0\\
0 & w_{10} & 0 & 0\\
0 & 0 & w_{01} & 0\\
0 & 0 & 0 & w_{00}
\end{bmatrix}
\times
\begin{bmatrix}
y_{11}\\
y_{10}\\
y_{01}\\
y_{00}
\end{bmatrix}\\
=&
\begin{bNiceMatrix}[%
  code-after={
    \tikz \node [highlight_red = (1-1) (1-4)] {} ;
    \tikz \node [highlight_blue = (2-1) (3-4)] {} ;
  }]
w_{11} & w_{10} & w_{01} & w_{00}\\
w_{11} & w_{10} & 0 & 0\\
w_{11} & 0 & w_{01} & 0
\end{bNiceMatrix}
\times
\begin{bmatrix}
y_{11}\\
y_{10}\\
y_{01}\\
y_{00}
\end{bmatrix}\\
=&
\begin{bNiceMatrix}[%
  code-after={
    \tikz \node [highlight_red = (1-1)] {} ;
    \tikz \node [highlight_blue = (2-1) (3-1)] {} ;
  }]
\sum{w_{ij}y_{ij}}\\
w_{11}y_{11} + w_{10}y_{10}\\
w_{11}y_{11} + w_{01}y_{01}
\end{bNiceMatrix}
\textnormal{.}
\end{align*}
This formulation, however, presupposes that all of the data points that share the same \((i, j)\) coordinates in the binary interpretable representation have the same target value (i.e., black-box prediction) \(y_{ij}\). %
To allow multiple copies of the same instance with different target values, we generalise this result by going back to \(\Theta\), which is the solution to the classic OLS. %
This approach is valid since weighted OLS for which the weights represent the count of each unique data point is equivalent to classic OLS for a data set whose instances are duplicated according to the counts given by the corresponding weights.%

Let us denote \(f: \mathcal{X} \rightarrow \mathcal{Y}\) as the black-box model and \(\IR: \mathcal{X} \rightarrow \mathcal{X}^\star\) as the transformation function from tabular data \(\mathcal{X}\) into their binary interpretable representation \(\mathcal{X}^\star\). %
Let us further define \(\mathcal{W}_{i j} = \{x \in \mathbf{X} : \IR(x) = (i, j)\}\) as the set of all the data points that are assigned to the same hyper-rectangle \((i, j)\) in the binary interpretable representation, and \(\mathcal{W} = \mathbf{X}\) as the set of all the data points. %
Now, recall that \(w_{ij}\) is the count of data points whose binary interpretable representation is \((i, j)\), therefore \(|\mathcal{W}_{ij}| = w_{ij}\) and \(|\mathcal{W}| = \sum{w_{ij}}\). %
Without loss of generality, we can reformulate the \(\mathbf{X}^T\mathbf{W}\mathbf{y}\) part of the \(\Theta_{\mathbf{W}}\) equation as \(\mathbf{X}^T\mathbf{y}\), which allows us to sum over all of the unique black-box predictions of instances assigned to particular hyper-rectangles:%
\begin{align*}
\mathbf{X}^T\mathbf{y} =& %
\begin{bNiceMatrix}[%
  code-after={
    \tikz \node [highlight_red = (1-1)] {} ;
    \tikz \node [highlight_blue = (2-1) (3-1)] {} ;
  }]
\sum{w_{ij}y_{ij}}\\
w_{11}y_{11} + w_{10}y_{10}\\
w_{11}y_{11} + w_{01}y_{01}
\end{bNiceMatrix}
=
\begin{bNiceMatrix}[%
  code-after={
    \tikz \node [highlight_red = (1-1)] {} ;
    \tikz \node [highlight_blue = (2-1) (3-1)] {} ;
  }]
\sum_{i \in \mathcal{W}}{y_i}\\
\sum_{i \in \mathcal{W}_{11}\cup\mathcal{W}_{10}} y_i\\
\sum_{i \in \mathcal{W}_{11}\cup\mathcal{W}_{01}} y_i
\end{bNiceMatrix}
\textnormal{.}
\end{align*}
This step allows us to relax the assumption of duplicated target values \(y_{ij}\), hence avoid imposing restrictions on the type of the model under investigation (probabilistic, crisp or regressor) and whether the binary representation has full fidelity with respect to the black box.%
\footnote{Note that in the original formulation of \(\mathbf{X}^T\mathbf{W}\mathbf{y}\) instances located within each hyper-rectangle determined by the underlying interpretable representation are assumed to share the same black-box prediction. %
This constraint makes the original weighted solution incompatible with regressors and probabilistic classifiers that are in need of explainability; %
for crisp classifiers, on the other hand, it implies that the binarised data space respects the black-box decision surface, i.e., it achieves full fidelity.%
}%

Finally, to better understand the meaning of influence-based explanations, we reformulate the sum of black-box predictions as their average:%
\begin{align*}
\mathbf{X}^T\mathbf{y} =& %
\begin{bNiceMatrix}[%
  code-after={
    \tikz \node [highlight_red = (1-1)] {} ;
    \tikz \node [highlight_blue = (2-1) (3-1)] {} ;
  }]
\sum_{i \in \mathcal{W}}{y_i}\\
\sum_{i \in \mathcal{W}_{11}\cup\mathcal{W}_{10}} y_i\\
\sum_{i \in \mathcal{W}_{11}\cup\mathcal{W}_{01}} y_i
\end{bNiceMatrix}
=
\begin{bNiceMatrix}[%
  code-after={
    \tikz \node [highlight_red = (1-1)] {} ;
    \tikz \node [highlight_blue = (2-1) (3-1)] {} ;
  }]
\sum_{i \in \mathcal{W}}{y_i} / \sum{w_{ij}} \times \sum{w_{ij}}\\
\sum_{i \in \mathcal{W}_{11}\cup\mathcal{W}_{10}} y_i / (w_{11}+w_{10}) \times (w_{11}+w_{10})\\
\sum_{i \in \mathcal{W}_{11}\cup\mathcal{W}_{01}} y_i / (w_{11}+w_{01}) \times (w_{11}+w_{01})%
\end{bNiceMatrix}\\
=&
\begin{bNiceMatrix}[%
  code-after={
    \tikz \node [highlight_red = (1-1)] {} ;
    \tikz \node [highlight_blue = (2-1) (3-1)] {} ;
  }]
\widebar{y}_{\mathcal{W}} \times \sum{w_{ij}}\\
\widebar{y}_{\mathcal{W}_{11}\cup\mathcal{W}_{10}} \times (w_{11}+w_{10})\\
\widebar{y}_{\mathcal{W}_{11}\cup\mathcal{W}_{01}} \times (w_{11}+w_{01})%
\end{bNiceMatrix}\\
=&%
\begin{bmatrix}
1 & 0 & 0\\
0 & 1 & 0\\
0 & 0 & 1
\end{bmatrix}
\times%
\begin{bNiceMatrix}[%
  code-after={
    \tikz \node [highlight_red = (1-1)] {} ;
    \tikz \node [highlight_blue = (2-1) (3-1)] {} ;
  }]
\widebar{y}_{\mathcal{W}} \times \sum{w_{ij}}\\
\widebar{y}_{\mathcal{W}_{11}\cup\mathcal{W}_{10}} \times (w_{11}+w_{10})\\
\widebar{y}_{\mathcal{W}_{11}\cup\mathcal{W}_{01}} \times (w_{11}+w_{01})%
\end{bNiceMatrix}\\
=&%
\begin{bNiceMatrix}[%
  columns-width = auto,
  code-after={
    \tikz \node [highlight_blue = (2-2) (3-3)] {} ;
    \tikz\path [fill=red!15, blend mode = multiply, name suffix = -medium, rounded corners = 0.5 mm]
    ($(1-1.north east)+(0pt,1pt)$)
    -- ($(1-3.north east)+(1pt,1pt)$)
    -- ($(1-3.south east)+(1pt,-1pt)$)
    -- ($(1-1.south east)+(1pt,-1pt)$)
    -- ($(3-1.south east)+(1pt,-1pt)$)
    -- ($(3-1.south west)+(-1pt,-1pt)$)
    -- ($(1-1.north west)+(-1pt,1pt)$)
    -- cycle ;
  }]
\sum{w_{ij}} & 0 & 0\\
0 & w_{11}+w_{10} & 0\\
0 & 0 & w_{11}+w_{01}
\end{bNiceMatrix}
\times%
\begin{bNiceMatrix}[%
  code-after={
    \tikz \node [highlight_red = (1-1)] {} ;
    \tikz \node [highlight_blue = (2-1) (3-1)] {} ;
  }]
\widebar{y}_{\mathcal{W}}\\
\widebar{y}_{\mathcal{W}_{11}\cup\mathcal{W}_{10}}\\
\widebar{y}_{\mathcal{W}_{11}\cup\mathcal{W}_{01}}
\end{bNiceMatrix}
\textnormal{,}
\end{align*}
and combine this result with \(\mathbf{X}^T\mathbf{W}\mathbf{X}\):%
\begin{align*}
&
{\begin{bNiceMatrix}[%
  columns-width = auto,
  code-after={
    \tikz \node [highlight_blue = (2-2) (3-3)] {} ;
    \tikz\path [fill=red!15, blend mode = multiply, name suffix = -medium, rounded corners = 0.5 mm]
    ($(1-1.north east)+(0pt,1pt)$)
    -- ($(1-3.north east)+(1pt,1pt)$)
    -- ($(1-3.south east)+(1pt,-1pt)$)
    -- ($(1-1.south east)+(1pt,-1pt)$)
    -- ($(3-1.south east)+(1pt,-1pt)$)
    -- ($(3-1.south west)+(-1pt,-1pt)$)
    -- ($(1-1.north west)+(-1pt,1pt)$)
    -- cycle ;
  }]
\sum{w_{ij}}         & w_{11} + w_{10} & w_{11} + w_{01}\\
w_{11} + w_{10} & w_{11} + w_{10} & w_{11}         \\
w_{11} + w_{01} & w_{11}          & w_{11} + w_{01}
\end{bNiceMatrix}}^{-1}
\times%
\begin{bNiceMatrix}[%
  columns-width = auto,
  code-after={
    \tikz \node [highlight_blue = (2-2) (3-3)] {} ;
    \tikz\path [fill=red!15, blend mode = multiply, name suffix = -medium, rounded corners = 0.5 mm]
    ($(1-1.north east)+(0pt,1pt)$)
    -- ($(1-3.north east)+(1pt,1pt)$)
    -- ($(1-3.south east)+(1pt,-1pt)$)
    -- ($(1-1.south east)+(1pt,-1pt)$)
    -- ($(3-1.south east)+(1pt,-1pt)$)
    -- ($(3-1.south west)+(-1pt,-1pt)$)
    -- ($(1-1.north west)+(-1pt,1pt)$)
    -- cycle ;
  }]
\sum{w_{ij}} & 0 & 0\\
0 & w_{11}+w_{10} & 0\\
0 & 0 & w_{11}+w_{01}
\end{bNiceMatrix}\\
&\times
\begin{bNiceMatrix}[%
  code-after={
    \tikz \node [highlight_red = (1-1)] {} ;
    \tikz \node [highlight_blue = (2-1) (3-1)] {} ;
  }]
\widebar{y}_{\mathcal{W}}\\
\widebar{y}_{\mathcal{W}_{11}\cup\mathcal{W}_{10}}\\
\widebar{y}_{\mathcal{W}_{11}\cup\mathcal{W}_{01}}
\end{bNiceMatrix}\\
=&
{\begin{bNiceMatrix}[%
  columns-width = auto,
  code-after={
    \tikz \node [highlight_blue = (2-2) (3-3)] {} ;
    \tikz\path [fill=red!15, blend mode = multiply, name suffix = -medium, rounded corners = 0.5 mm]
    ($(1-1.north east)+(0pt,1pt)$)
    -- ($(1-3.north east)+(1pt,1pt)$)
    -- ($(1-3.south east)+(1pt,-1pt)$)
    -- ($(1-1.south east)+(1pt,-1pt)$)
    -- ($(3-1.south east)+(1pt,-1pt)$)
    -- ($(3-1.south west)+(-1pt,-1pt)$)
    -- ($(1-1.north west)+(-1pt,1pt)$)
    -- cycle ;
  }]
\sum{w_{ij}}         & w_{11} + w_{10} & w_{11} + w_{01}\\
w_{11} + w_{10} & w_{11} + w_{10} & w_{11}         \\
w_{11} + w_{01} & w_{11}          & w_{11} + w_{01}
\end{bNiceMatrix}}^{-1}
\times%
{\begin{bNiceMatrix}[%
  columns-width = auto,
  code-after={
    \tikz \node [highlight_blue = (2-2) (3-3)] {} ;
    \tikz\path [fill=red!15, blend mode = multiply, name suffix = -medium, rounded corners = 0.5 mm]
    ($(1-1.north east)+(0pt,1pt)$)
    -- ($(1-3.north east)+(1pt,1pt)$)
    -- ($(1-3.south east)+(1pt,-1pt)$)
    -- ($(1-1.south east)+(1pt,-1pt)$)
    -- ($(3-1.south east)+(1pt,-1pt)$)
    -- ($(3-1.south west)+(-1pt,-1pt)$)
    -- ($(1-1.north west)+(-1pt,1pt)$)
    -- cycle ;
  }]
\frac{1}{\sum{w_{ij}}} & 0 & 0\\
0 & \frac{1}{w_{11}+w_{10}} & 0\\
0 & 0 & \frac{1}{w_{11}+w_{01}}
\end{bNiceMatrix}}^{-1}\\
&\times
\begin{bNiceMatrix}[%
  code-after={
    \tikz \node [highlight_red = (1-1)] {} ;
    \tikz \node [highlight_blue = (2-1) (3-1)] {} ;
  }]
\widebar{y}_{\mathcal{W}}\\
\widebar{y}_{\mathcal{W}_{11}\cup\mathcal{W}_{10}}\\
\widebar{y}_{\mathcal{W}_{11}\cup\mathcal{W}_{01}}
\end{bNiceMatrix}\\
=&\left(
\begin{bNiceMatrix}[%
  columns-width = auto,
  code-after={
    \tikz \node [highlight_blue = (2-2) (3-3)] {} ;
    \tikz\path [fill=red!15, blend mode = multiply, name suffix = -medium, rounded corners = 0.5 mm]
    ($(1-1.north east)+(0pt,1pt)$)
    -- ($(1-3.north east)+(1pt,1pt)$)
    -- ($(1-3.south east)+(1pt,-1pt)$)
    -- ($(1-1.south east)+(1pt,-1pt)$)
    -- ($(3-1.south east)+(1pt,-1pt)$)
    -- ($(3-1.south west)+(-1pt,-1pt)$)
    -- ($(1-1.north west)+(-1pt,1pt)$)
    -- cycle ;
  }]
\frac{1}{\sum{w_{ij}}} & 0 & 0\\
0 & \frac{1}{w_{11}+w_{10}} & 0\\
0 & 0 & \frac{1}{w_{11}+w_{01}}
\end{bNiceMatrix}
\times
\begin{bNiceMatrix}[%
  columns-width = auto,
  code-after={
    \tikz \node [highlight_blue = (2-2) (3-3)] {} ;
    \tikz\path [fill=red!15, blend mode = multiply, name suffix = -medium, rounded corners = 0.5 mm]
    ($(1-1.north east)+(0pt,1pt)$)
    -- ($(1-3.north east)+(1pt,1pt)$)
    -- ($(1-3.south east)+(1pt,-1pt)$)
    -- ($(1-1.south east)+(1pt,-1pt)$)
    -- ($(3-1.south east)+(1pt,-1pt)$)
    -- ($(3-1.south west)+(-1pt,-1pt)$)
    -- ($(1-1.north west)+(-1pt,1pt)$)
    -- cycle ;
  }]
\sum{w_{ij}}         & w_{11} + w_{10} & w_{11} + w_{01}\\
w_{11} + w_{10} & w_{11} + w_{10} & w_{11}         \\
w_{11} + w_{01} & w_{11}          & w_{11} + w_{01}
\end{bNiceMatrix}
\right)^{-1}\\
&\times%
\begin{bNiceMatrix}[%
  code-after={
    \tikz \node [highlight_red = (1-1)] {} ;
    \tikz \node [highlight_blue = (2-1) (3-1)] {} ;
  }]
\widebar{y}_{\mathcal{W}}\\
\widebar{y}_{\mathcal{W}_{11}\cup\mathcal{W}_{10}}\\
\widebar{y}_{\mathcal{W}_{11}\cup\mathcal{W}_{01}}
\end{bNiceMatrix}\\
=&
{\begin{bNiceMatrix}[%
  columns-width = auto,
  code-after={
    \tikz \node [highlight_blue = (2-2) (3-3)] {} ;
    \tikz\path [fill=red!15, blend mode = multiply, name suffix = -medium, rounded corners = 0.5 mm]
    ($(1-1.north east)+(0pt,1pt)$)
    -- ($(1-3.north east)+(1pt,1pt)$)
    -- ($(1-3.south east)+(1pt,-1pt)$)
    -- ($(1-1.south east)+(1pt,-1pt)$)
    -- ($(3-1.south east)+(1pt,-1pt)$)
    -- ($(3-1.south west)+(-1pt,-1pt)$)
    -- ($(1-1.north west)+(-1pt,1pt)$)
    -- cycle ;
  }]
1 & \frac{w_{11} + w_{10}}{\sum{w_{ij}}} & \frac{w_{11} + w_{01}}{\sum{w_{ij}}}\\
1 & 1 & \frac{w_{11}}{w_{11} + w_{10}}\\
1 & \frac{w_{11}}{w_{11} + w_{01}} & 1
\end{bNiceMatrix}}^{-1}
\times
\begin{bNiceMatrix}[%
  code-after={
    \tikz \node [highlight_red = (1-1)] {} ;
    \tikz \node [highlight_blue = (2-1) (3-1)] {} ;
  }]
\widebar{y}_{\mathcal{W}}\\
\widebar{y}_{\mathcal{W}_{11}\cup\mathcal{W}_{10}}\\
\widebar{y}_{\mathcal{W}_{11}\cup\mathcal{W}_{01}}
\end{bNiceMatrix}
\textnormal{.}
\end{align*}
This formulation demonstrates an unexpected role of the:%
\begin{enumerate}
  \item number of data points sampled in each hyper-rectangle on the resulting explanations (magnitudes of concept influence); and%
  \item irrelevance of the feature partitions other than the ones determining the hyper-rectangles that directly enclose the explained instance.%
\end{enumerate}
A more detailed discussion of the interpretation and significance of this result can be found in Section~\ref{sec:ols}.%

\end{document}